\documentclass[10pt,letterpaper]{article}
\usepackage[T1]{fontenc}
\usepackage{textcomp}
\usepackage{multirow}
\usepackage{mathtools}
\usepackage{arydshln}
\usepackage{authblk}
\usepackage[hyphenbreaks]{breakurl}
\usepackage[labelformat=empty,position=bottom]{subfig}
\usepackage[normal,flushleft]{threeparttable}
\usepackage{nameref}
\usepackage{pgfplots}
\pgfplotsset{compat=1.13}
\usetikzlibrary{patterns}
\usetikzlibrary{pgfplots.colormaps}

\pgfplotsset{
  /pgfplots/area1 legend/.style={
    legend image code/.code={%
    \draw[#1] (0cm,-0.05cm) rectangle (0.3cm,0.05cm);
    } 
}
}

\usepackage{setspace}
\DeclarePairedDelimiter\floor{\lfloor}{\rfloor}

\newcommand{\FDGPETscore}{FDG-PET DAT score}
\newcommand{\FDGPETscoreShort}{FPDS}

\newcommand{\TauAbetaRatio}{t-tau/A$\beta_{1\mbox{-}42}$}

\newcommand{\myTODO}[1]{}

\makeatletter
\newcommand\resetstackedplots[1]{
\makeatletter
\pgfplots@stacked@isfirstplottrue
\makeatother
\addplot [forget plot,draw=none] coordinates{#1};
}
\makeatother

\definecolor{darkgreen}{rgb}{0.0,0.5,0.0}
\definecolor{darkred}{rgb}{0.55,0.0,0.0}
\definecolor{sNC}{rgb}{0.01, 0.72, 0.24}  
\definecolor{pMCI}{rgb}{1, 0.95, 0.1}   
\definecolor{sMCI}{rgb}{0, 0.7, 1}  
\definecolor{pNC}{rgb}{0.93, 0.51, 0.93}  
\definecolor{eDAT}{rgb}{0.5, 0.5, 0} 
\definecolor{uNC}{rgb}{0.94, 0.19, 0} 
\definecolor{sDAT}{rgb}{0, 0.2, 0.7}    

\title{Development and validation of a novel dementia of Alzheimer`s type (DAT) score based on metabolism FDG-PET imaging}

\author[a]{Karteek Popuri}
\author[a]{Rakesh Balachandar}
\author[a]{Kathryn Alpert}
\author[a]{Donghuan Lu} 
\author[a]{Mahadev Bhalla}
\author[c]{Ian Mackenzie}
\author[c]{Robin Ging-Yuek Hsiung}
\author[b]{Lei Wang}
\author[a]{Mirza Faisal Beg}
\author[ ]{the Alzheimer\textquotesingle s Disease Neuroimaging Initiative\thanks{Data used in preparation of this article were obtained from the Alzheimer\textquotesingle s Disease Neuroimaging Initiative (ADNI) database (\protect\url{http://adni.loni.usc.edu}). As such, the investigators within the ADNI contributed to the design and implementation of ADNI and/or provided data but did not participate in analysis or writing of this report. A complete listing of ADNI investigators can be found at: \protect\url{http://adni.loni.usc.edu/wp-content/uploads/how_to_apply/ADNI_Acknowledgement_List.pdf}}}

\affil[a]{Simon Fraser University}
\affil[b]{Northwestern University}
\affil[c]{University of British Columbia}
\begin{document}

\maketitle

\begin{abstract}
Fluorodeoxyglucose positron emission tomography (FDG-PET) imaging based 3D topographic brain glucose metabolism patterns from normal controls (NC) and individuals with dementia of Alzheimer's type (DAT) are used to train a novel multi-scale ensemble classification model. This ensemble model outputs a \FDGPETscore\ (\FDGPETscoreShort) between 0 and 1 denoting the probability of a subject to be clinically diagnosed with DAT based on their metabolism profile. A novel 7 group image stratification scheme is devised that groups images not only based on their associated clinical diagnosis but also on past and future trajectories of the clinical diagnoses, yielding a more continuous representation of the different stages of DAT spectrum that mimics a real-world clinical setting. The potential for using \FDGPETscoreShort\ as a DAT biomarker was validated on a large number of FDG-PET images ($N$=2984) obtained from the  Alzheimer\textquotesingle s Disease Neuroimaging Initiative (ADNI) database taken across the proposed stratification, and a good classification AUC (area under the curve) of 0.78 was achieved in distinguishing between images belonging to subjects on a DAT trajectory and those images taken from subjects not progressing to a DAT diagnosis. Further, the \FDGPETscoreShort\ biomarker achieved state-of-the-art performance on the mild cognitive impairment (MCI) to DAT conversion prediction task with an AUC of 0.81, 0.80, 0.77 for the 2, 3, 5 years to conversion windows respectively.
\end{abstract}

\begin{table*}
\centering
\caption{Novel stratification of ADNI images and associated demographic, clinical \& biomarker details. The stratification was based on two criteria, clinical diagnosis of subjects at the time of FDG-PET image acquisition and their longitudinal clinical progression. Each image is assigned a membership of the form `prefix{Group}', where `Group' is the clinical diagnosis at imaging visit, and `prefix' signals past or future clinical diagnoses. For e.g., an image is designated as pNC if the subject was assigned a NC diagnosis at that particular imaging visit, but the subject converts to DAT at a future timepoint. The eDAT images are associated with the diagnosis of DAT, but the subject had received NC or MCI status during previous ADNI visits (conversion within ADNI window). Whereas, the sDAT images belong to subjects with a consistent clinical diagnosis of DAT throughout the ADNI study window, hence these individuals have progressed to DAT prior to their ADNI recruitment.}
\begin{threeparttable}
{\scriptsize \setlength{\tabcolsep}{0.022in}\begin{tabular}{*{8}{l}}
\hline
&&Clinical&&&&&\\
&&diagnosis&&&&&\\
Dementia & Group &at& Clinical & $N^{3}$ & Age & MMSE & CSF\\
trajectory & name &imaging& progression & {[}images{]} & {[}Years{]} & {[}Max. 30{]} & {[}\TauAbetaRatio{]}\\
\hline
DAT- & sNC:stable NC$^5$ & NC & \textbf{NC} $\rightarrow$ \textbf{NC} & 753 & 75.44 (5.95) & 29.08 (1.17) & 0.37 (0.26)\\
DAT- & uNC:unstable NC & NC & \textbf{NC} $\rightarrow$ MCI & 110 & 78.93 (4.91) & 29.05 (1.13) & 0.47 (0.32)\\
DAT- & sMCI:stable MCI & MCI & NC $\rightarrow$ \textbf{MCI} \textit{or} \textbf{MCI} $\rightarrow$ \textbf{MCI} & 881 & 75.02 (7.77) & 27.86 (1.95) & 0.55 (0.47)\\
\hline
DAT+ & pNC:progressive NC & NC & \textbf{NC} $\rightarrow$ MCI $\rightarrow$ DAT & 58 & 78.20 (4.43) & 28.90 (1.29) & 0.59 (0.27)\\
DAT+ & pMCI:progressive MCI & MCI & NC $\rightarrow$ \textbf{MCI} $\rightarrow$ DAT \textit{or} \textbf{MCI} $\rightarrow$ DAT & 486 & 74.87 (7.12) & 26.77 (2.06) & 0.88 (0.52)\\
DAT+ & eDAT:early DAT & DAT & NC $\rightarrow$ MCI $\rightarrow$ \textbf{DAT} \textit{or} MCI $\rightarrow$ \textbf{DAT} & 232 & 76.59 (6.77) & 22.25 (4.51) & 0.94 (0.62)\\
DAT+ & sDAT:stable DAT$^6$ & DAT & \textbf{DAT} $\rightarrow$ \textbf{DAT} & 464 & 75.80 (7.49) & 22.02 (3.64) & 1.03 (0.58)\\
\hline
\end{tabular}
}
{\scriptsize
\begin{tablenotes}[flushleft]
     \item[$1$] NC: normal controls, MCI: mild cognitive impairment, DAT: dementia of Alzheimer's type
     \item[{\color{white} $1$}] MMSE: mini mental state examination, CSF: cerebrospinal fluid, t-tau: total tau, A$\beta_{1\mbox{-}42}$: beta amyloid $1\mbox{-}42$
     \item[$2$] DAT+: On DAT trajectory, i.e., at some point in time, these subjects will be clinically diagnosed as DAT
     \item[{\color{white} $2$}] DAT-: not on the DAT trajectory and will not get a DAT diagnosis in the ADNI window
     \item[$3$] A total of $2984$ FDG-PET images were taken from $1298$ subjects
     \item[{\color{white} $3$}] Number of subjects corresponding to images in each of the groups:
     \item[{\color{white} $3$}] sNC (360), uNC (52), sMCI (431), pNC (18), pMCI (205), eDAT (133), sDAT (238)
     \item[{\color{white} $3$}] Number of subjects with images across multiple groups:
     \item[{\color{white} $3$}] uNC \& sMCI (18), pNC \& pMCI (7), pNC \& eDAT (6), pMCI \& eDAT (110), pNC \& pMCI \& eDAT (2)
     \item[$4$] The mean (standard deviation) age, MMSE score and CSF measure values within each group are given
     \item[{\color{white} $4$}] CSF measures were only available for a subset of images in each of the groups:
     \item[{\color{white} $4$}] sNC (384), uNC (48), sMCI (470), pNC (24), pMCI (205), eDAT (66), sDAT (230)
     \item[$5$] baseline sNC: $N$=360, Age: 73.81 (6.07), MMSE: 29.05 (1.22), CSF: 0.36 (0.25)
     \item[{\color{white} $5$}] follow-up sNC: $N$=393, Age: 76.93 (5.44), MMSE: 29.11 (1.11), CSF: 0.39 (0.28)
    \item[$6$] baseline sDAT: $N$=238, Age: 74.93 (7.87), MMSE: 23.22 (2.13), CSF: 1.02 (0.58)
    \item[{\color{white} $6$}] follow-up sDAT: $N$=226, Age: 76.71 (6.97), MMSE: 20.76 (4.40), CSF: 1.06 (0.58)
\end{tablenotes}
}
\end{threeparttable}
\label{tab:group_definitions_demographics}
\end{table*}
\section{Introduction}
Alzheimer's disease (AD) is a neurodegenerative disorder characterized by the presence of AD pathology (ADP) such as aberrant deposition of amyloid beta (A$\beta$) proteins, and the appearance of neurofibrillary tangles of tau proteins. The initial symptom of AD is cognitive impairment notably in the memory domain, that gradually involves other domains leading to a clinical diagnosis of dementia of Alzheimer's type (DAT). Patients with DAT progressively succumb to severe stages of dementia, requiring complete assistance for daily activities. DAT is the most common form of dementia, affecting 1 in 9 people over the age of 65 years \cite{AlzAssoc2015FactsFigures} and as many as 1 in 3 people over the age of 85 \cite{Hebert2013}. As of 2015, there were an estimated 46.8 million dementia afflicted growing to reach 131.5 million in 2050 \cite{Prince2016}, projecting a very sizeable burden on healthcare systems and caregivers worldwide. This impending public health crisis due to rising DAT cases has prompted drug-development efforts to find treatments for AD that can reduce the severity of ADP or remove it altogether \cite{Cummings2014,Godyn2016}. However, the success of such treatments ultimately depends on the ability to diagnose DAT as early as possible before irreversible brain damage occurs. Therefore, in recent years there has been a considerable push towards developing robust biomarkers useful for diagnosing DAT in clinical practice \cite{Weiner2017}.

Fluorodeoxyglucose positron emission tomography (FDG-PET) is a minimally invasive neuroimaging technique to quantify the glucose metabolism in the brain which indirectly measures the underlying neuronal activity \cite{Mosconi2010FdgPetReview}. As metabolic disruptions are hypothesized to precede the appearance of cognitive symptoms in AD \cite{Jack2013}, FDG-PET imaging presents itself as an attractive tool for investigating the metabolism changes triggered by ADP across the entire DAT spectrum, ranging from the presymptomatic phase to the mild cognitive impairment (MCI) stage followed by dementia. Our aim in this work is to develop an automatic method that can aid in the interpretation of the 3D topographic metabolism patterns encoded in FDG-PET images for the purpose of DAT diagnosis. To this end, we devised a supervised machine learning framework that takes as input a FDG-PET image of subject and outputs a continuous value between 0 and 1 termed as the \FDGPETscore\ (\FDGPETscoreShort), which indicates the probability of the subject's metabolism profile to be belonging to the DAT trajectory, i.e., how likely is the subject to be clinically diagnosed with DAT.

One of the main contributions of our work is the introduction of a novel approach for stratifying the imaging data used in the development and validation of the proposed \FDGPETscoreShort\ methodology. Most commonly, imaging biomarker studies employ a 3 group stratification, where the clinical diagnostic labels of NC, MCI and DAT assigned at the time of image acquisition are directly used for grouping the imaging data \cite{Rathore2017}. In contrast, here we present a stratification scheme that groups images based not only on their associated clinical diagnosis but also on past and future clinical diagnoses. Our novel stratification is able to more faithfully represent the different diagnostic trajectories observed in a real-world clinical setting when compared to the stratification depending only on the diagnosis at a single timepoint. For instance, based on our stratification, we can distinguish among NC images that stay NC (stable NC, sNC) from those that convert to MCI (unstable NC, uNC), and from those that convert to DAT (progressive NC, pNC). A similar delineation is also induced among the MCI and DAT images using our stratification scheme. An important contribution in this paper is the design of a novel multi-scale ensemble classification model for the proposed \FDGPETscoreShort\ computation. The ensemble model consists of several individual classifiers trained on features extracted from the FDG-PET image at multiple scales. The probability predictions from each of these individual classifiers regarding the association of the given FDG-PET image with a DAT trajectory are fused together to obtain a more robust final \FDGPETscoreShort\ prediction. Another noteworthy contribution of our work is the exhaustive and comprehensive statistical evaluation approach used to validate the \FDGPETscoreShort\ predictions. First, the training model fit was evaluated and then a pseudo-independent test sample consisting of follow-up images corresponding to the baseline training data was used to obtain a more accurate estimate of the ensemble model’s generalization error. Finally, the predictive performance of the \FDGPETscoreShort\ biomarker was evaluated on a large completely independent validation set of images taken from different stages of the DAT spectrum demonstrating a strong generalization potential of the reported results. To the best of our knowledge, ours is the largest FDG-PET based imaging biomarker study reported till date.

\section{Methods}
\label{secn:methods}
\subsection{Study participants}
Data used in the preparation of this article was obtained from the ADNI database (\url{adni.loni.usc.edu}). The ADNI was launched in 2003 as a public-private partnership, led by principal investigator Michael W. Weiner, MD. The primary goal of ADNI has been to test whether serial MRI, PET, other biological markers, and clinical and neuropsychological assessment can be combined to measure the progression of MCI and early Alzheimer’s disease. Till date, ADNI has involved 1887 subjects and assessed over one or more visits. Clinical diagnosis received by these subjects, can be broadly categorized among one of NC, MCI and DAT. Detailed description of the ADNI recruitment procedure, image acquisition protocols and diagnostic criteria can be found at \url{www.adni-info.org}.

\subsection{Novel database stratification}
We devised a novel stratification scheme to distinguish within the NC, MCI and DAT groups based on past and future clinical diagnosis received by the individual. Each of these three groups were further divided into subgroups based on the diagnoses received during their follow-up. The subgroups are named according to the convention `prefix{Group}', where `Group' is the clinical diagnosis obtained during the imaging visit, and `prefix' signifies the past or the future clinical diagnoses of the same individual. Images associated with clinical diagnosis of NC, and a consistent diagnoses of NC during the entire ADNI study period are termed as the stable NC (sNC) group. Images associated with clinical diagnosis of NC, but convert to MCI in the future visits are termed as unstable NC (uNC). Images associated with a clinical diagnosis of NC and convert to DAT in their future visits are termed as progressive NC (pNC). Similarly, images associated with MCI are subgrouped as stable MCI (sMCI) and progressive MCI (pMCI) based on persistent MCI diagnosis and conversion to DAT diagnosis respectively in their subsequent followup. Images with a clinical diagnosis of DAT who joined ADNI at the DAT stage, i.e., they converted to clinical diagnosis of DAT prior to ADNI recruitment, and remained DAT for the future ADNI visits are termed as stable DAT (sDAT). Images with a clinical diagnosis of DAT, with the recent past ADNI clinical diagnosis of either NC or MCI, i.e., they converted to DAT within the ADNI visits are termed as early DAT (eDAT). Note that a past or future clinical diagnosis visit may or may not include neuroimaging, but the past or future clinical diagnosis enables an enriched staging of each image given the evolution of clinical diagnosis.

The proposed stratification provides key advantage, offers subgroups namely pNC, pMCI, eDAT and sDAT, that represent various stages of DAT trajectory. The pNC subgroup is the earliest, the sDAT subgroup is the most advanced and the pMCI and eDAT subgroups are in-between these extremes along the DAT spectrum.  These are denoted as the DAT+ class of images indicating their trajectory towards DAT. The subjects in the sNC, uNC and sMCI subgroups do not include a followup clinical diagnosis of DAT during the ADNI window; so although there is the possibility that post-ADNI these could progress to a clinical diagnosis of DAT, for the purposes of analysis in this paper, these subgroups are considered to not be on the DAT+ trajectory, hence denoted as DAT-.

\subsection{MRI processing}
Pre-processing of the 3D structural MPRAGE T1-weighted MRI images from ADNI included standard intensity normalization to remove image geometry distortions arising from gradient non-linearity, B1 calibrations to correct for image intensity non-uniformities and N3 histogram peak sharpening (\url{http://adni.loni.usc.edu/methods/mri-analysis/mri-pre-processing}). The pre-processed images were segmented into the gray matter (GM), white matter (WM) and cerebrospinal fluid (CSF) tissue regions \cite{Dale1999FreesurferCorticalReconstruction} using the Freesurfer software package (\url{https://surfer.nmr.mgh.harvard.edu}). A rigorous quality control procedure was employed to manually identify and correct any errors in the automated tissue segmentations following Freesurfer's troubleshooting guidelines. Subsequently, the GM tissue region was parcellated into 85 different anatomical ROIs using Freesurfer's cortical \cite{desikan2006automated} and subcortical \cite{Fischl2002Freesurfer} labeling pipelines.

\subsection{FDG-PET processing}
The ADNI FDG-PET images used in this study were pre-processed using a series of steps to mitigate inter-scanner variability and obtain FDG-PET data with a uniform spatial resolution and intensity range for further analysis (\url{http://adni.loni.usc.edu/methods/PET-analysis/pre-processing}). Briefly, the original raw FDG-PET frames were co-registered and averaged to obtain a single FDG-PET image, which was then mapped from its native space to a standard $160\times160\times96$ image grid with $1.5\times1.5\times1.5$ mm$^3$ voxels. After standardizing the spatial resolution and orientation, the intensity range of the FDG-PET image was normalized such that average intensity of all the foreground voxels in the image was exactly equal to one. The intensity normalized images were then filtered using scanner-specific filter functions to obtain FDG-PET data at a uniform smoothing level of isotropic $8$ mm full width at half maximum (FWHM) Gaussian kernel.

\subsection{Multi-scale patch-wise FDG-PET SUVR features}
In order to better localize the average regional glucose metabolism signal, each of the $85$ GM ROIs obtained using Freesurfer were further subdivided into smaller volumetric sub-regions or \textit{patches}. Our previously proposed adaptive surface patch generation method \cite{Raamana2015ThickNetFusion_NBA}, which is based on $k$-means clustering, was applied to the 3D image domain to obtain a patch-wise parcellation of the GM ROIs. Instead of subdividing each GM ROI into a fixed number of patches, the number of patches per ROI were adaptively determined using the patch size parameter ($m$), denoting the number of voxels in each patch. This achieves a patch density (patches in ROI/voxels in ROI) that is uniform ($\approx \frac{1}{m}$) throughout the image domain, which is desirable, as it leads to a compact yet rich description of the entire GM tissue region. The scale-space theory framework \cite{witkin1984scale} argues for storing the signal at multiple scales in the absence of a-priori knowledge regarding the appropriate scale at which to analyze the signal. Motivated by this scale-space idea, we generated $16$ different levels of patch-wise parcellations, $m$ = $\{100,150,200,250,300,350,400,450,500,$ $1000,1500,2000,3000,4000,5000,10000\}$ to obtain a fine to coarse multi-scale representation of the GM region for capturing the regional glucose metabolism signals at different scales. We note that the patch-wise parcellations were initially generated on the standard MNI ICBM 152 non-linear average T1 template \cite{grabner2006symmetric} (\url{http://nist.mni.mcgill.ca/?p=858}) and then were propagated to each of the target MRI images in our dataset using the large deformation diffeomorphic metric mapping (LDDMM) non-rigid registration \cite{Beg2005lddmm}. This template-based parcellation approach ensures a one-to-one correspondence between the target image patches, which is required for the construction of a valid multi-scale FDG-PET feature space in the next step. 

The FDG-PET images were co-registered with their respective MRI images using the inter-modal linear registration facility \cite{jenkinson2002improved} available as part of the FSL-FLIRT program (\url{https://fsl.fmrib.ox.ac.uk/fsl/fslwiki/FLIRT}). The quality of the co-registration was visually checked and the detected failures were corrected by re-running FSL-FLIRT with a narrower rotation angle search range parameter to avoid getting trapped in local minima. The estimated $12$ degrees of freedom (DOF) mapping was used to transfer the patch-wise parcellations from the MRI domain onto the FDG-PET domain. The mean FDG-PET image intensity value in each of the mapped patches was used to calculate the patch-wise standardized uptake value ratios (SUVRs) as, the mean intensity in a given patch divided by the mean intensity in the brainstem, chosen as the reference ROI. This resulted in a total of $M = 17$ (including the original Freesurfer parcellation) patch-wise FDG-PET SUVR feature vectors that encoded the multi-scale regional glucose metabolism information derived from a given target FDG-PET image.

\subsection{\FDGPETscore\ computation via supervised ensemble learning}
A supervised classification framework following the well established ensemble learning paradigm was used to calculate the proposed \FDGPETscore\ from the multi-scale patch-wise SUVR feature vectors. The main idea behind ensemble based supervised classification is to combine several individually trained classifiers together to obtain a single, more robust classification model \cite{dietterich2000ensemble}. Accordingly, in the proposed framework, classifiers were trained separately on each of the individual multi-scale feature vector spaces to construct a classifier ensemble. Then, a fusion of the multiple predictions from individual classifiers in the ensemble was performed, yielding the ensemble model estimate about the probability of the input multi-scale feature vectors belonging to the DAT+ trajectory. This probabilistic prediction output by the ensemble classification framework was taken to be the proposed \FDGPETscore.

The training samples corresponding to the DAT- and DAT+ classes needed for building the ensemble classification model were given by the baseline sNC ($N$=$360$) and sDAT ($N$=$238$) images respectively (Table \ref{tab:group_definitions_demographics}, footnotes 5 and 6). The proposed $M$ multi-scale patch-wise FDG-PET SUVR feature vectors were extracted from all the training samples. To prevent over-fitting of the ensemble model to the chosen training sample set, the subagging approach \cite{buhlmann2003bagging} was employed to randomly generate $F = 100$ subsets of training samples. The random sampling was performed using a sampling ratio of $\gamma = 0.8$ in a stratified manner to avoid class imbalance, ensuring an equal number of samples from both the DAT- and DAT+ classes, i.e., $N_{train} = 2 \times \floor{0.8 \times 238} = 380$ samples in each of the $F$ training subsets. An ensemble of $M \times F$ probabilistic kernel \cite{Damoulas2008ProteinFoldRecog} classifiers were individually trained on each of the $M$ feature spaces using the $F$ different training subsets. The classifier training was preceded by a t-statistic based feature selection step to identify the $k = \floor{N_{train}/10} = 38$ most discriminative features within the feature vector and also to address the ``curse of dimensionality'' issue \cite{Raamana2015ThickNetFusion_NBA}. Each of the $M \times F = 1700$ trained probabilistic kernel classifiers output a continuous scalar $p_{i} \in [0\;1], i = \{1,\ldots M \times F\}$, that denotes the probability of an input feature vector belonging to the DAT+ class ($1-p_{i}$ being the DAT- class membership probability). The \FDGPETscore\  is then simply defined as the mean of the DAT+ class probability predictions obtained from each of the $M \times F$ classifiers. 

In summary, given an unseen ``test'' sample containing a FDG-PET/MRI image pair, we first extract the $M$ multi-scale patch-wise SUVR features vectors from the images, and then reduce the dimensionality of each of these feature vectors by retaining only the $k$ most discriminative features that were identified during the training phase. The pruned feature vectors are fed to the previously trained $M \times F$ classifier ensemble to obtain $M \times F$ probability predictions regarding the DAT+ class membership, which are then averaged to obtain the \FDGPETscore\  corresponding to the given test sample.

\section{Results}
Our study dataset consisted of $2984$ FDG-PET images (with corresponding structural MRI images), belonging to $1294$ ADNI subjects, who have undergone imaging and clinical evaluations at one or more longitudinal time points. The images were stratified into one of the $7$ study groups based on the clinical diagnosis received at the time of image acquisition and the clinical diagnosis received previously and/or during subsequent follow-up time points (Table \ref{tab:group_definitions_demographics}). 

In the proposed stratification scheme, we distinguish among the images that have a clinical diagnosis of NC (sNC, uNC, pNC) at the imaging visit. Within this NC group, there are NC that will stay NC, i.e., stable NC (sNC, $N$=$753$ images), convert to MCI, i.e., unstable NC (uNC, $N$=$110$ images) or convert to DAT, i.e., progressive NC (pNC, $N$=$58$ images), and hence even though all are NC, the images are treated as distinct subgroups of the NC group given their future divergent evolution of clinical diagnosis. In a similar fashion, we distinguish among the images with clinical diagnosis of MCI as consisting of those who will continue to stay MCI, i.e., stable MCI (sMCI, $N$=$881$ images) throughout ADNI, or convert to AD, i.e., progressive MCI (pMCI, $N$=$486$ images) at a future visit. Finally, we distinguish among those images that have an associated clinical diagnosis of DAT. Those DAT that had a previous clinical diagnosis of NC or MCI, i.e., joined ADNI as either NC or MCI and converted to DAT during ADNI are denoted as the early DAT group (eDAT, $N$=$232$ images) given their recent conversion, whereas those that joined ADNI with a clinical diagnosis of DAT and hence their conversion was prior to their ADNI recruitment and remained DAT throughout the ADNI window are designated as the stable DAT (sDAT, $N$=$464$ images). There are 110 individuals with FDG-PET images at both the pMCI and the eDAT stages, i.e., these individuals underwent conversion from MCI to DAT during the ADNI window and this conversion was sampled with neuroimaging.

\subsection{Demographic, clinical \& biomarker values across groups} The $7$ stratified image sets were compared \begin{table}
    \centering
    \caption{The $p$-values corresponding to the significance of the pairwise group differences in the age, MMSE score and CSF \TauAbetaRatio\ measure values among the $7$ stratified groups. The $t$-test or Wilcoxon ranksum test was used depending on if the data followed a normal distribution or not. The cases where the group mean values were significantly ($p$<$0.001$) different are highlighted in bold and the cases where data followed a normal distribution are underlined.}
    \setlength{\tabcolsep}{0.02in}\begin{tabular}{c c c c | c c c c}
\hline
Groups & Age & MMSE & CSF & Groups & Age & MMSE & CSF\\
\hline
sNC-uNC & \textbf{\underline{<0.0001}} & 0.5276 & 0.0046 & sMCI-pNC & 0.0028 & \textbf{<0.0001} & 0.0555\\
sNC-sMCI & 0.8034 & \textbf{<0.0001} & \textbf{<0.0001} & sMCI-pMCI & 0.6312 & \textbf{<0.0001} & \textbf{<0.0001}\\
sNC-pNC & \textbf{\underline{<0.0001}} & 0.3760 & \textbf{<0.0001} & sMCI-eDAT & 0.0149 & \textbf{<0.0001} & \textbf{<0.0001}\\
sNC-pMCI & 0.4997 & \textbf{<0.0001} & \textbf{<0.0001} & sMCI-sDAT & 0.1029 & \textbf{<0.0001} & \textbf{<0.0001}\\
sNC-eDAT & \underline{0.0211} & \textbf{<0.0001} & \textbf{<0.0001} & pNC-pMCI & \textbf{0.0005} & \textbf{<0.0001} & 0.0029\\
sNC-sDAT & 0.0932 & \textbf{<0.0001} & \textbf{<0.0001} & pNC-eDAT & \underline{0.0290} & \textbf{<0.0001} & 0.0055\\
uNC-sMCI & \textbf{<0.0001} & \textbf{<0.0001} & 0.5432 & pNC-sDAT & 0.0181 & \textbf{<0.0001} & \textbf{<0.0001}\\
uNC-pNC & \underline{0.3340} & 0.6900 & 0.0170 & pMCI-eDAT & 0.0046 & \textbf{<0.0001} & 0.8320\\
uNC-pMCI & \textbf{<0.0001} & \textbf{<0.0001} & \textbf{<0.0001} & pMCI-sDAT & 0.0424 & \textbf{<0.0001} & 0.0047\\
uNC-eDAT & \textbf{\underline{0.0003}} & \textbf{<0.0001} & \textbf{<0.0001} & eDAT-sDAT & 0.2709 & 0.0945 & 0.1072\\
uNC-sDAT & \textbf{<0.0001} & \textbf{<0.0001} & \textbf{<0.0001} &  &  &  &  \\
\end{tabular}
    \label{tab:pvalue_demographics}
\end{table}
for group-level differences in their associated age, mini mental state exam (MMSE) score and CSF \TauAbetaRatio\ measure (ratio of total tau to beta amyloid $1\mbox{-}42$) values. Pairwise significance testing of the group mean value differences was performed between all the groups, using the $t$-test in the case of normally distributed data and the Wilcoxon ranksum test for the non-parametric data distribution case. The $p$-values obtained from each of the pairwise significance tests are reported in Table \ref{tab:pvalue_demographics}. The statisical significance threshold was set at $p$<$0.001$. The mean age was observed to be statistically similar across all the groups except for the uNC and pNC groups which exhibited significantly higher ages. The mean MMSE scores were significantly higher among the sNC, uNC and pNC groups when compared to either the sMCI and pMCI groups or the eDAT and sDAT groups. The DAT- (sNC, uNC, sMCI) groups had significantly lower mean CSF \TauAbetaRatio\ measures when compared to the DAT+ (pNC, pMCI, eDAT, sDAT) groups apart from the two cases where pNC showed statistically similar CSF \TauAbetaRatio\ measures compared to uNC and sMCI respectively.
\subsection{Automatic salient ROI selection for \FDGPETscoreShort\ computation} 
\begin{table}
\centering
\caption{Most discriminative ROIs chosen by the ensemble classification model. The ROIs are listed in descending order of their total (left and right averaged) selection frequency. Note that only ROIs with a non-zero selection frequency (selected at least once) are shown.}
\setlength{\tabcolsep}{0.02in}\begin{tabular}{c c|c c}
\hline
ROI & Frequency ($\%$)& ROI & Frequency ($\%$)\\
name& [Left | Right]& name & [Left | Right]\\
\hline
isthmuscingulate & 100.00 | 99.65 & fusiform & 24.29 | 0.53\\
precuneus & 100.00 | 83.88 & medialorbitofrontal & 12.76 | 10.29\\
inferiortemporal & 99.82 | 83.35 & superiorfrontal & 14.29 | 5.94\\
posteriorcingulate & 96.12 | 85.06 & superiortemporal & 11.94 | 5.24\\
middletemporal & 99.35 | 80.71 & lateralorbitofrontal & 12.18 | 2.24\\
inferiorparietal & 99.18 | 64.94 & superiorparietal & 11.41 | 3.00\\
supramarginal & 67.41 | 26.06 & parsopercularis & 9.88 | 1.06\\
entorhinal & 57.94 | 32.53 & temporalpole & 9.35 | 0.18\\
hippocampus & 47.82 | 32.00 & rostralanteriorcingulate & 5.18 | 0.00\\
bankssts & 27.76 | 15.82 & frontalpole & 0.82 | 0.82\\
rostralmiddlefrontal & 24.94 | 17.18 & caudate & 0.71 | 0.00\\
amygdala & 22.18 | 17.29 & parstriangularis & 0.35 | 0.00\\
parahippocampal & 28.00 | 10.06 & parsorbitalis & 0.18 | 0.00\\
caudalmiddlefrontal & 22.76 | 13.18 & & \\
\end{tabular}
\label{tab:discrimROIs}
\end{table}
The feature selection phase of the ensemble classification model training identified several ROIs that contained strong discriminatory FDG uptake information useful for separating the DAT- and DAT+ classes. Specifically, each of the individual $1700$ classifiers in the ensemble model automatically selected a set of $38$ most discriminative ROIs from which the multi-scale patch-wise FDG-PET SUVR features were taken and used to compute the \FDGPETscoreShort. In Table \ref{tab:discrimROIs}, selection frequencies of the ROIs chosen by the classifier ensemble are listed. The selection frequency of a ROI is defined as the fraction of the classifiers in the ensemble that chose the particular ROI. Interestingly, ROIs from the left hemisphere exhibited much higher selection frequencies compared to the corresponding right hemisphere ROIs. Further, the cortical ROIs had far greater selection frequencies than the subcortical ROIs. In particular, the isthmus and posterior parts of the cingulate gyrus, the precuneus and the inferior and middle temporal gyri had very high (>90\%) total (left and right averaged) selection frequencies.

\subsection{\FDGPETscoreShort\ distribution among training (sNC and sDAT) groups} 
\begin{figure}
\centering
\caption{\FDGPETscoreShort\ distribution among the sNC and sDAT images and classification performance obtained in assigning images to either the DAT- or DAT+ trajectory using a 0.5 \FDGPETscoreShort\ threshold. The top row presents the out-of-bag predictions on the baseline images, which were used for training the ensemble model. The bottom row shows ensemble model predictions on the follow-up subgroup. The follow-up images were not part of training and hence were considered as unseen test samples for the purpose of \FDGPETscoreShort\ computation. The (number of images : mean \FDGPETscoreShort) is shown for each subgroup. Balanced accuracy is the mean of the sensitivity and specificity measures.}
\subfloat{\begin{tikzpicture}
\begin{axis}[ybar stacked,xmin=0.0000,xmax=1.0100,ymin=0.0000,ymax=0.6500,width=0.7\linewidth,height=0.42\linewidth,
xtick={0,0.1,0.2,0.3,0.4,0.5,0.6,0.7,0.8,0.9,1},
xticklabels={0,0.1,0.2,0.3,0.4,0.5,0.6,0.7,0.8,0.9,1},
ytick={0,0.2,0.4,0.6},
yticklabels={0,0.2,0.4,0.6},
extra x ticks={0.5},extra x tick labels = {},
extra y ticks={0.2,0.4,0.6},extra y tick labels = {},
extra tick style = {grid=major},
xlabel={\FDGPETscoreShort},xlabel shift = -10pt,
ylabel={Fraction of images},ylabel shift = -10pt,
xtick pos = left,xtick align = outside,
ytick pos = left,ytick align = outside,
area legend,bar width=0.0300,
x tick label style={font=\small},y tick label style={font=\small},x label style={font=\small},y label style={font=\small},
legend columns = 1,legend style={/tikz/every even column/.append style={column sep=0.18cm}},
legend entries={baseline sNC ($360:0.1645$),baseline sDAT ($238:0.7962$)},legend style={anchor=south},legend style={at={(0.5,1.01)}},
legend style={font=\small,cells={align=center}},legend image post style={scale=0.95}]
\addplot[bar shift=-.018,draw=sNC,fill=sNC,opacity=1,pattern=,pattern color=sNC] table[x=x,y=y1] {training_error-stacked-1.txt};
\resetstackedplots{(0.05000,0.000000) (0.15000,0.000000) (0.25000,0.000000) (0.35000,0.000000) (0.45000,0.000000) (0.55000,0.000000) (0.65000,0.000000) (0.75000,0.000000) (0.85000,0.000000) (0.95000,0.000000)}
\addplot[bar shift=.018,draw=sDAT,fill=sDAT,opacity=1,pattern=,pattern color=sDAT] table[x=x,y=y1] {training_error-stacked-2.txt};
\resetstackedplots{(0.05000,0.000000) (0.15000,0.000000) (0.25000,0.000000) (0.35000,0.000000) (0.45000,0.000000) (0.55000,0.000000) (0.65000,0.000000) (0.75000,0.000000) (0.85000,0.000000) (0.95000,0.000000)}
\node at (axis cs:0.45,0.48) {\setlength{\tabcolsep}{0.02in}\begin{tabular}{rc}AUC:&$0.9537$\\Accuracy:&$0.8980$\\Specificity:&$0.9167$\\Sensitivity:&$0.8697$\\Balanced accuracy:&$0.8932$\\\end{tabular}};
\end{axis}
\end{tikzpicture}
}\\
\subfloat{\begin{tikzpicture}
\begin{axis}[ybar stacked,xmin=0.0000,xmax=1.0100,ymin=0.0000,ymax=0.7000,width=0.7\linewidth,height=0.42\linewidth,
xtick={0,0.1,0.2,0.3,0.4,0.5,0.6,0.7,0.8,0.9,1.0},
xticklabels={0,0.1,0.2,0.3,0.4,0.5,0.6,0.7,0.8,0.9,1.0},
ytick={0,0.2,0.4,0.6},
yticklabels={0,0.2,0.4,0.6},
extra x ticks={0.5},extra x tick labels = {},
extra y ticks={0.2,0.4,0.6},extra y tick labels = {},
extra tick style = {grid=major},
xlabel={\FDGPETscoreShort},xlabel shift = -10pt,
ylabel={Fraction of images},ylabel shift = -10pt,
xtick pos = left,xtick align = outside,
ytick pos = left,ytick align = outside,
area legend,bar width=0.0300,
x tick label style={font=\small},y tick label style={font=\small},x label style={font=\small},y label style={font=\small},
legend columns = 1,legend style={/tikz/every even column/.append style={column sep=0.18cm}},
legend entries={follow-up sNC ($393:0.2210$),follow-up sDAT ($226:0.8866$)},legend style={anchor=south},legend style={at={(0.5,1.01)}},
legend style={font=\small,cells={align=center}},legend image post style={scale=0.95}]
\addplot[bar shift=-.018,draw=sNC,fill=sNC,opacity=1,pattern=,pattern color=sNC] table[x=x,y=y1] {testing_error-stacked-1.txt};
\resetstackedplots{(0.05000,0.000000) (0.15000,0.000000) (0.25000,0.000000) (0.35000,0.000000) (0.45000,0.000000) (0.55000,0.000000) (0.65000,0.000000) (0.75000,0.000000) (0.85000,0.000000) (0.95000,0.000000)}
\addplot[bar shift=.018,draw=sDAT,fill=sDAT,opacity=1,pattern=,pattern color=sDAT] table[x=x,y=y1] {testing_error-stacked-2.txt};
\resetstackedplots{(0.05000,0.000000) (0.15000,0.000000) (0.25000,0.000000) (0.35000,0.000000) (0.45000,0.000000) (0.55000,0.000000) (0.65000,0.000000) (0.75000,0.000000) (0.85000,0.000000) (0.95000,0.000000)}
\node at (axis cs:0.5,0.52) {\setlength{\tabcolsep}{0.02in}\begin{tabular}{rc}AUC:&$0.9789$\\Accuracy:&$0.8869$\\Specificity:&$0.8499$\\Sensitivity:&$0.9513$\\Balanced accuracy:&$0.9006$\\\end{tabular}};
\end{axis}
\end{tikzpicture}
}
\label{fig:fpds_of_sNC_sDAT}
\end{figure}
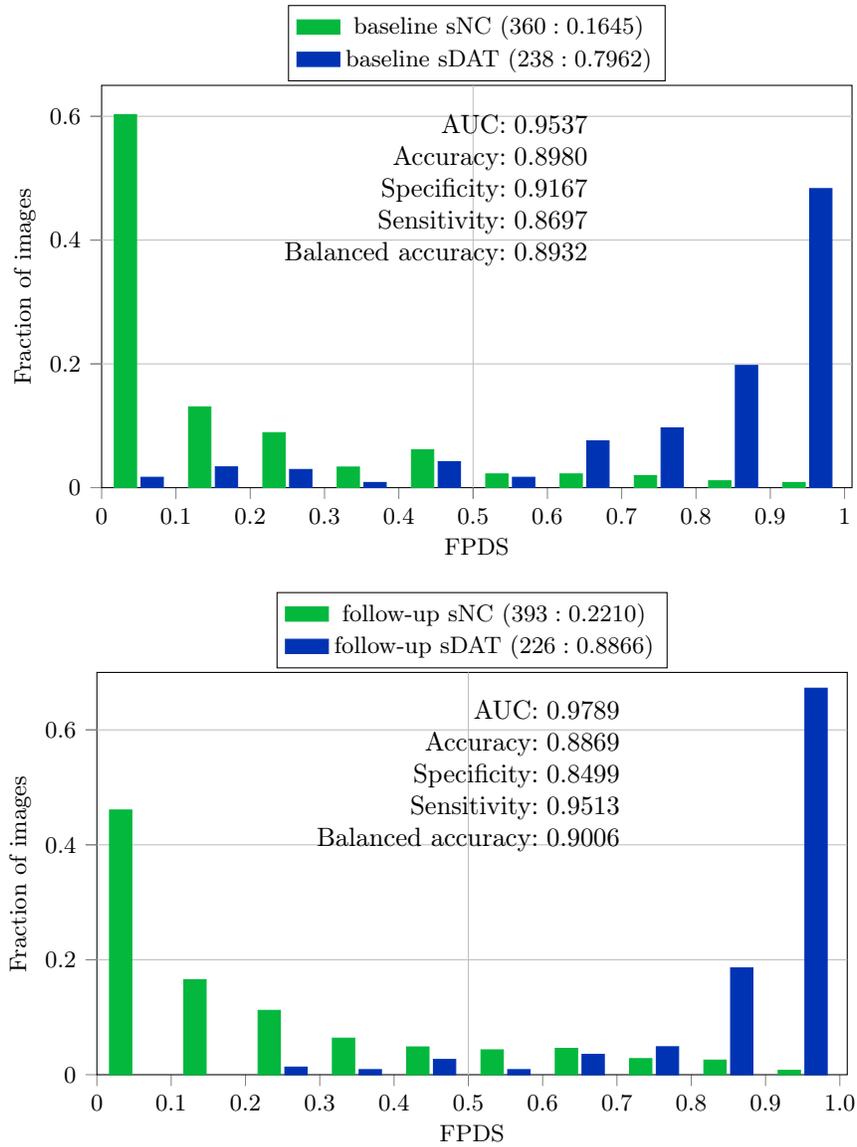
In Figure \ref{fig:fpds_of_sNC_sDAT}, the distribution of \FDGPETscoreShort\ values among the baseline and follow-up images from the sNC and sDAT groups are shown. As the baseline images were used for training the ensemble model, the \FDGPETscoreShort\ values for the baseline images were determined via the out-of-bag prediction approach to avoid biased estimates. In this approach, the \FDGPETscoreShort\ for a given baseline image was computed by only fusing predictions from classifiers in the ensemble that did not have the given baseline image as part of their subagging training subset. The follow-up images were not involved in the ensemble model training, so they were treated as unseen test samples and their \FDGPETscoreShort\ values were computed using the standard approach of fusing predictions from all the classifiers in the ensemble. It can be seen from Figure \ref{fig:fpds_of_sNC_sDAT} that the \FDGPETscoreShort\ distributions of the sNC and sDAT groups are very well separated with an excellent (>0.95) area under the curve (AUC) of the receiver operating characteristic (ROC) in both the baseline and follow-up image cases. Moreover, high specificities and sensitivities ($\sim$ 0.90 balanced accuracies) were achieved when using a \FDGPETscoreShort\ threshold of 0.5 to classify the baseline and follow-up images as belonging to either the DAT- or the DAT+ trajectory.

\subsection{\FDGPETscoreShort\ distribution among the validation image groups} 
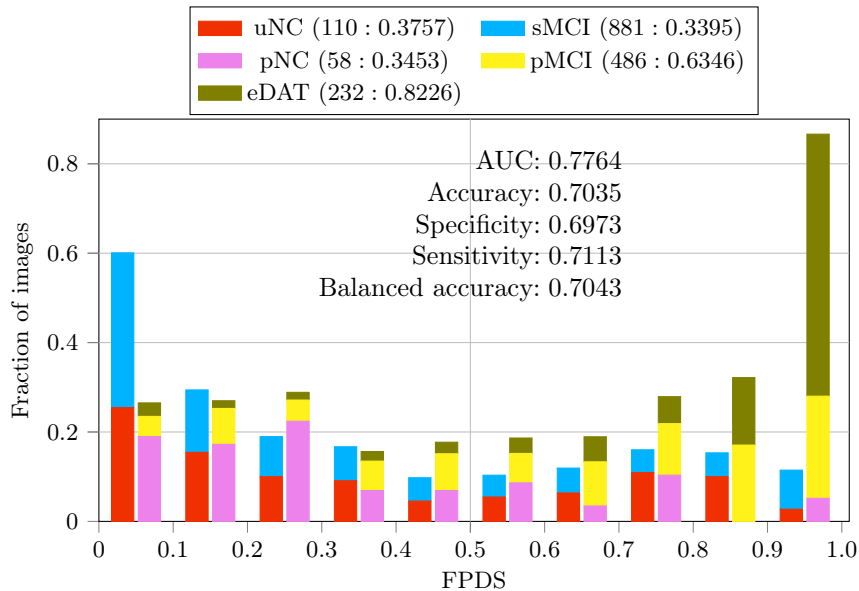
\begin{figure}
\centering
\caption{The \FDGPETscoreShort\ distribution among validation image groups and the classification performance obtained in determining dementia trajectories (DAT- or DAT+) for these images using a 0.5 \FDGPETscoreShort\ threshold. The \FDGPETscoreShort\ histograms corresponding to the groups on the DAT- (uNC, sMCI) and the DAT+ (pNC, pMCI, eDAT) trajectories are stacked together respectively. The (number of images : mean \FDGPETscoreShort) is shown for each group. Balanced accuracy is mean of sensitivity and specificity.}
\begin{tikzpicture}
\begin{axis}[ybar stacked,xmin=0.0000,xmax=1.0100,ymin=0.0000,ymax=0.9000,width=0.7\linewidth,height=0.42\linewidth,
xtick={0,0.1,0.2,0.3,0.4,0.5,0.6,0.7,0.8,0.9,1.0},
xticklabels={0,0.1,0.2,0.3,0.4,0.5,0.6,0.7,0.8,0.9,1.0},
ytick={0,0.2,0.4,0.6,0.8},
yticklabels={0,0.2,0.4,0.6,0.8},
extra x ticks={0.5},extra x tick labels = {},
extra y ticks={0.2,0.4,0.6,0.8},extra y tick labels = {},
extra tick style = {grid=major},
xlabel={\FDGPETscoreShort},xlabel shift = -10pt,
ylabel={Fraction of images},ylabel shift = -10pt,
xtick pos = left,xtick align = outside,
ytick pos = left,ytick align = outside,
area legend,bar width=0.0300,
x tick label style={font=\small},y tick label style={font=\small},x label style={font=\small},y label style={font=\small},
legend columns = 2,legend style={/tikz/every even column/.append style={column sep=0.18cm}},
legend entries={uNC ($110:0.3757$),sMCI ($881:0.3395$),pNC ($58:0.3453$),pMCI ($486:0.6346$),eDAT ($232:0.8226$)},legend style={anchor=south},legend style={at={(0.5,1.01)}},
legend style={font=\small,cells={align=center}},legend image post style={scale=0.95}]
\addplot[bar shift=-.018,draw=uNC,fill=uNC,opacity=1,pattern=,pattern color=uNC] table[x=x,y=y1] {validation_all-stacked-1.txt};
\addplot[bar shift=-.018,draw=sMCI,fill=sMCI,opacity=1,pattern=,pattern color=sMCI] table[x=x,y=y2] {validation_all-stacked-1.txt};
\resetstackedplots{(0.05000,0.000000) (0.15000,0.000000) (0.25000,0.000000) (0.35000,0.000000) (0.45000,0.000000) (0.55000,0.000000) (0.65000,0.000000) (0.75000,0.000000) (0.85000,0.000000) (0.95000,0.000000)}
\addplot[bar shift=.018,draw=pNC,fill=pNC,opacity=1,pattern=,pattern color=pNC] table[x=x,y=y1] {validation_all-stacked-2.txt};
\addplot[bar shift=.018,draw=pMCI,fill=pMCI,opacity=1,pattern=,pattern color=pMCI] table[x=x,y=y2] {validation_all-stacked-2.txt};
\addplot[bar shift=.018,draw=eDAT,fill=eDAT,opacity=1,pattern=,pattern color=eDAT] table[x=x,y=y3] {validation_all-stacked-2.txt};
\resetstackedplots{(0.05000,0.000000) (0.15000,0.000000) (0.25000,0.000000) (0.35000,0.000000) (0.45000,0.000000) (0.55000,0.000000) (0.65000,0.000000) (0.75000,0.000000) (0.85000,0.000000) (0.95000,0.000000)}
\node at (axis cs:0.5,0.66) {\setlength{\tabcolsep}{0.02in}\begin{tabular}{rc}AUC:&$0.7764$\\Accuracy:&$0.7035$\\Specificity:&$0.6973$\\Sensitivity:&$0.7113$\\Balanced accuracy:&$0.7043$\\\end{tabular}};
\end{axis}
\end{tikzpicture}
\label{fig:fpds_of_uNC_sMCI_pNC_pMCI_eDAT}
\end{figure}
Imaging data from the uNC and sMCI groups that belong to the DAT- trajectory, along with images from the pNC, pMCI and eDAT groups that are on the DAT+ trajectory constituted the independent validation set used for evaluating the proposed ensemble model framework for \FDGPETscoreShort\ computation. In Figure \ref{fig:fpds_of_uNC_sMCI_pNC_pMCI_eDAT}, \FDGPETscoreShort\ distributions across these independent validation image groups are shown. In general, the mean \FDGPETscoreShort\ values among the DAT- trajectory groups (<0.4) were much lower compared to the \FDGPETscoreShort\ group means across the DAT+ trajectory groups (>0.6), except for the pNC group which had a mean \FDGPETscoreShort\ value of 0.35 which was similar to that of the DAT- groups. It should however also be noted that the pNC group contained far fewer images ($N$=58) in comparison to the other groups. Overall, there was a good degree of separation between the DAT- and DAT+ \FDGPETscoreShort\ distributions resulting in an AUC of 0.78. Further, this separability translated into a balanced accuracy of 0.70 when the images were classified into either the DAT- or the DAT+ trajectory using a 0.5 \FDGPETscoreShort\ threshold.

\subsection{\FDGPETscoreShort\ trend across age ranges in validation image groups} 
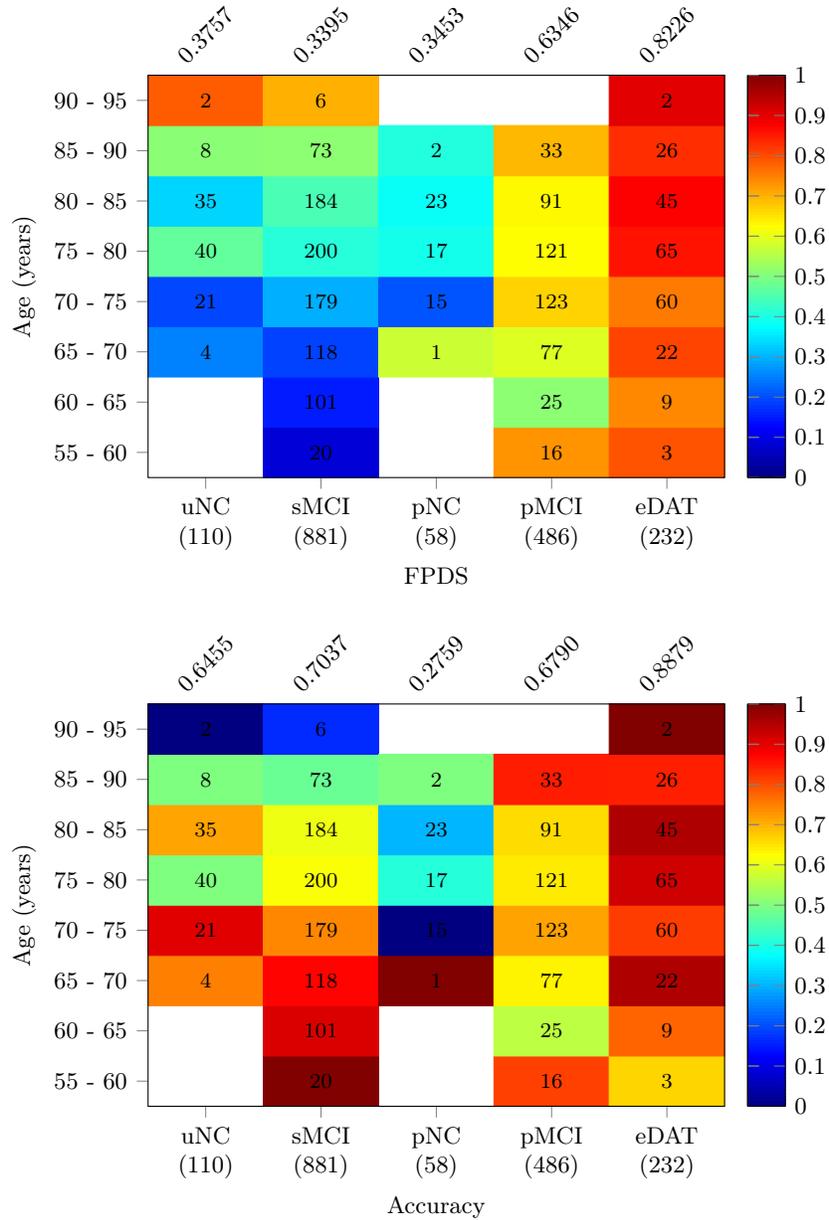
\begin{figure}
\centering
\caption{Age-based analysis of \FDGPETscoreShort\ score: heat map plots showcasing the trend of mean \FDGPETscoreShort\ (top) and classification accuracy (bottom) obtained across different age ranges within each of the validation image groups. The classification accuracies were calculated using a 0.5 \FDGPETscoreShort\ threshold. The number of images in a (image group,age range) is printed on the corresponding heat map cell, while the total number of images within a group is shown in parentheses under each column of the heat maps. The overall mean \FDGPETscoreShort\ and classification accuracy within a group are given above respective heat map columns.}
\subfloat{\begin{tikzpicture}
\begin{axis}[axis on top,clip=false,width=0.56\linewidth,height=0.42\linewidth,
colorbar sampled,
colorbar style={ytick={0,0.1,0.2,0.3,0.4,0.5,0.6,0.7,0.8,0.9,1},yticklabels={0,0.1,0.2,0.3,0.4,0.5,0.6,0.7,0.8,0.9,1},y tick label style={font=\small},samples=128
},
colormap/jet,
enlargelimits=false,
xtick={1,2,3,4,5},xticklabels={uNC\\(110),sMCI\\(881),pNC\\(58),pMCI\\(486),eDAT\\(232)},
ytick={57.5,62.5,67.5,72.5,77.5,82.5,87.5,92.5},yticklabels={55 - 60,60 - 65,65 - 70,70 - 75,75 - 80,80 - 85,85 - 90,90 - 95},
xtick pos = left,xtick align = outside,
ytick pos = left,ytick align = outside,
x tick label style={font=\small,align=center},
x label style={font=\small},xlabel shift = 0pt,
y tick label style={font=\small},
y label style={font=\small},ylabel shift = 0pt,
xlabel = {\FDGPETscoreShort},
ylabel = {Age (years)}]
\addplot [matrix plot*,mesh/cols=5,point meta=explicit,point meta min=0,point meta max=1] table [x=x,y=y,meta=z] {AGE_vs_FDGPETscore_ALL-hist_plot.txt};
\addplot [nodes near coords,nodes near coords style={font=\footnotesize},nodes near coords align={anchor=center},point meta=explicit symbolic,only marks,mark=none] table [x=x,y=y,meta=label] {AGE_vs_FDGPETscore_ALL-hist_plot.txt};
\node [rotate=45] at (axis cs:1,99) {\small 0.3757};
\node [rotate=45] at (axis cs:2,99) {\small 0.3395};
\node [rotate=45] at (axis cs:3,99) {\small 0.3453};
\node [rotate=45] at (axis cs:4,99) {\small 0.6346};
\node [rotate=45] at (axis cs:5,99) {\small 0.8226};
\end{axis}
\end{tikzpicture}}\\
\subfloat{\begin{tikzpicture}
\begin{axis}[axis on top,clip=false,width=0.56\linewidth,height=0.42\linewidth,
colorbar sampled,
colorbar style={ytick={0,0.1,0.2,0.3,0.4,0.5,0.6,0.7,0.8,0.9,1},yticklabels={0,0.1,0.2,0.3,0.4,0.5,0.6,0.7,0.8,0.9,1},y tick label style={font=\small},samples=128
},
colormap/jet,
enlargelimits=false,
xtick={1,2,3,4,5},xticklabels={uNC\\(110),sMCI\\(881),pNC\\(58),pMCI\\(486),eDAT\\(232)},
ytick={57.5,62.5,67.5,72.5,77.5,82.5,87.5,92.5},yticklabels={55 - 60,60 - 65,65 - 70,70 - 75,75 - 80,80 - 85,85 - 90,90 - 95},
xtick pos = left,xtick align = outside,
ytick pos = left,ytick align = outside,
x tick label style={font=\small,align=center},
x label style={font=\small},xlabel shift = 0pt,
y tick label style={font=\small},
y label style={font=\small},ylabel shift = 0pt,
xlabel = {Accuracy},
ylabel = {Age (years)}]
\addplot [matrix plot*,mesh/cols=5,point meta=explicit,point meta min=0,point meta max=1] table [x=x,y=y,meta=z] {AGE_vs_Accuracy_ALL-hist_plot.txt};
\addplot [nodes near coords,nodes near coords style={font=\footnotesize},nodes near coords align={anchor=center},point meta=explicit symbolic,only marks,mark=none] table [x=x,y=y,meta=label] {AGE_vs_Accuracy_ALL-hist_plot.txt};
\node [rotate=45] at (axis cs:1,99) {\small 0.6455};
\node [rotate=45] at (axis cs:2,99) {\small 0.7037};
\node [rotate=45] at (axis cs:3,99) {\small 0.2759};
\node [rotate=45] at (axis cs:4,99) {\small 0.6790};
\node [rotate=45] at (axis cs:5,99) {\small 0.8879};
\end{axis}
\end{tikzpicture}}
\label{fig:fpds_vs_AGE}
\end{figure}
The mean \FDGPETscoreShort\ values and classification accuracies (based on 0.5 \FDGPETscoreShort\ threshold) obtained from the validation image subsets taken across different age ranges within the uNC and sMCI groups (DAT- trajectory), and the pNC, pMCI and eDAT groups (DAT+ trajectory) are presented in Figure \ref{fig:fpds_vs_AGE}. The \FDGPETscoreShort\ means in the sMCI group gradually increased from less than 0.2 in the younger age ranges (55 - 70 years) to greater than 0.5 among the older age ranges (85 - 95 years). This wide and gradual variation of \FDGPETscoreShort\ values manifested as a steady decrease in the accuracy of identifying the sMCI images as DAT- from above 0.85 in the younger age ranges (55 - 70 years) to below 0.5 in the older age ranges (85 - 95 years). The eDAT group exhibited uniformly high \FDGPETscoreShort\ mean values across all the age ranges lying in a short interval of 0.74 - 0.9. Consequently, a majority of the eDAT images were correctly classified as DAT+, leading to a high overall accuracy of 0.89. In contrast to the sMCI and eDAT groups, no consistent age-related patterns of \FDGPETscoreShort\ mean values and classification accuracies were observed among the uNC, pNC and pMCI groups. The pMCI group displayed relatively high \FDGPETscoreShort\ mean values (>0.67) in three disjoint age ranges 55 - 60, 70 - 75 and 85 - 90 years, and accordingly the classification accuracies of 0.85, 0.71 and 0.81 respectively observed in these age ranges, were considerably higher than the overall pMCI group average of 0.68. Surprisingly, the pNC and uNC groups were found to have a similar \FDGPETscoreShort\ mean in the 70 - 75 years range, and further in the following 75 - 80 and 85 - 90 years ranges the pNC group had lower \FDGPETscoreShort\ means relative to the uNC group. This lead to mis-labeling of most pNC images as DAT-, yielding a very poor overall classification accuracy of 0.28.

\subsection{\FDGPETscoreShort\ versus time to conversion in progressive image groups} 
\begin{figure}
\centering
\caption{Heat maps showing variation of mean \FDGPETscoreShort\ (left) and classification accuracy (right) across different time to conversion ranges in the progressive image groups (pNC and pMCI). The time to conversion indicates the number of years from the image scan date to the first clinical diagnosis of DAT for the subject associated with the image. A \FDGPETscoreShort\ threshold of 0.5 was used to calculate the classification accuracies. The number of images in a (image group,time to conversion range) is printed on the corresponding heat map cell, while the total number of images within a group is shown in parentheses under each column of the heat maps. The overall mean \FDGPETscoreShort\ and classification accuracy within a group are given above respective heat map columns.}
\subfloat{\begin{tikzpicture}
\begin{axis}[axis on top,clip=false,width=0.294\linewidth,height=0.462\linewidth,
colormap/jet,
enlargelimits=false,
xtick={1,2},xticklabels={pNC\\(58),pMCI\\(486)},
ytick={0.5,1.5,2.5,3.5,4.5,5.5,6.5,7.5,8.5,9.5},yticklabels={0 - 1,1 - 2,2 - 3,3 - 4,4 - 5,5 - 6,6 - 7,7 - 8,8 - 9,9 - 10},
xtick pos = left,xtick align = outside,
ytick pos = left,ytick align = outside,
x tick label style={font=\small,align=center},
x label style={font=\small},xlabel shift = 0pt,
y tick label style={font=\small},
y label style={font=\small},ylabel shift = 0pt,
xlabel = {\FDGPETscoreShort},
ylabel = {Time to conversion (years)}]
\addplot [matrix plot*,mesh/cols=2,point meta=explicit,point meta min=0,point meta max=1] table [x=x,y=y,meta=z] {TimeToConversion_vs_FDGPETscore_progressive-hist_plot.txt};
\addplot [nodes near coords,nodes near coords style={font=\footnotesize},nodes near coords align={anchor=center},point meta=explicit symbolic,only marks,mark=none] table [x=x,y=y,meta=label] {TimeToConversion_vs_FDGPETscore_progressive-hist_plot.txt};
\node [rotate=45] at (axis cs:1,11) {\small 0.3453};
\node [rotate=45] at (axis cs:2,11) {\small 0.6346};
\end{axis}
\end{tikzpicture}}\qquad
\subfloat{\begin{tikzpicture}
\begin{axis}[axis on top,clip=false,width=0.294\linewidth,height=0.462\linewidth,
colorbar sampled,
colorbar style={ytick={0,0.1,0.2,0.3,0.4,0.5,0.6,0.7,0.8,0.9,1},yticklabels={0,0.1,0.2,0.3,0.4,0.5,0.6,0.7,0.8,0.9,1},y tick label style={font=\small},samples=128
},
colormap/jet,
enlargelimits=false,
xtick={1,2},xticklabels={pNC\\(58),pMCI\\(486)},
ytick=\empty,
xtick pos = left,xtick align = outside,
ytick pos = left,ytick align = outside,
x tick label style={font=\small,align=center},
x label style={font=\small},xlabel shift = 0pt,
y tick label style={font=\small},
y label style={font=\small},ylabel shift = 0pt,
xlabel = {Accuracy}]
\addplot [matrix plot*,mesh/cols=2,point meta=explicit,point meta min=0,point meta max=1] table [x=x,y=y,meta=z] {TimeToConversion_vs_Accuracy_progressive-hist_plot.txt};
\addplot [nodes near coords,nodes near coords style={font=\footnotesize},nodes near coords align={anchor=center},point meta=explicit symbolic,only marks,mark=none] table [x=x,y=y,meta=label] {TimeToConversion_vs_Accuracy_progressive-hist_plot.txt};
\node [rotate=45] at (axis cs:1,11) {\small 0.2759};
\node [rotate=45] at (axis cs:2,11) {\small 0.6790};
\end{axis}
\end{tikzpicture}}
\label{fig:fpds_vs_TimeToConversion}
\end{figure}
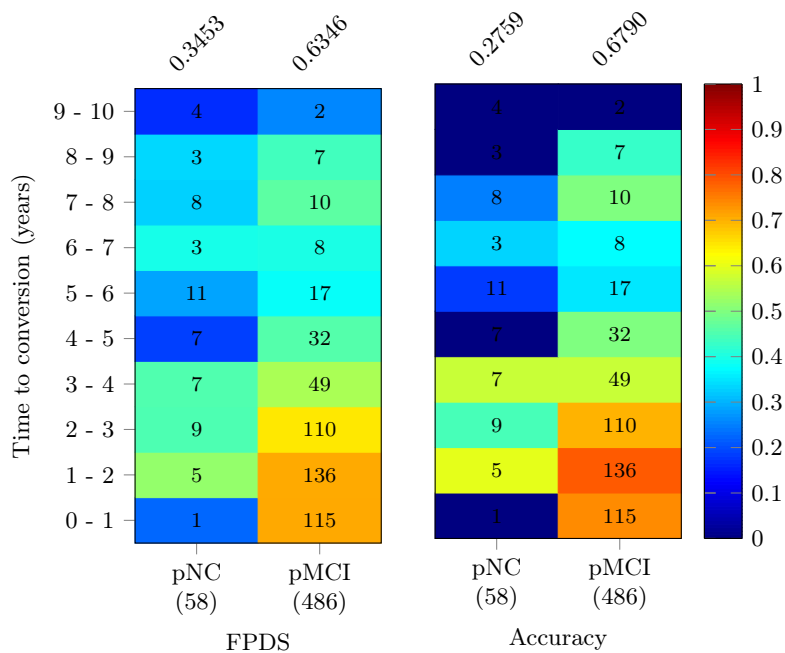
In Figure \ref{fig:fpds_vs_TimeToConversion}, the mean \FDGPETscoreShort\ values and classification accuracies (based on 0.5 \FDGPETscoreShort\ threshold) computed from image subsets taken across different ranges of time to conversion within the the pNC and pMCI groups are shown. The time to conversion is defined as the number of years from the image scan date to the earliest future timepoint at which the subject associated with the image was given a clinical diagnosis of DAT. The pMCI group exhibited relatively high mean \FDGPETscoreShort\ values (0.64 - 0.71) among the 0 - 3 years to conversion range. But, in the later time to conversion ranges, especially beyond the 4 years to conversion range, a considerable decrease (0.26 - 0.46) in the \FDGPETscoreShort\ means was observed. Therefore, for the pMCI group, good classification accuracies (0.7 - 0.78) were only observed in the 0 - 3 years to conversion range, past which the pMCI images were frequently misclassified as DAT-, reducing the overall accuracy to 0.68. The pNC group showed low \FDGPETscoreShort\ mean values (0.17 - 0.52) across all the time to conversion ranges, leading to incorrect labeling of more than 72\% of the pNC images as DAT- (0.28 overall classification accuracy).  

\subsection{Correlation between \FDGPETscoreShort\ and CSF \TauAbetaRatio} 
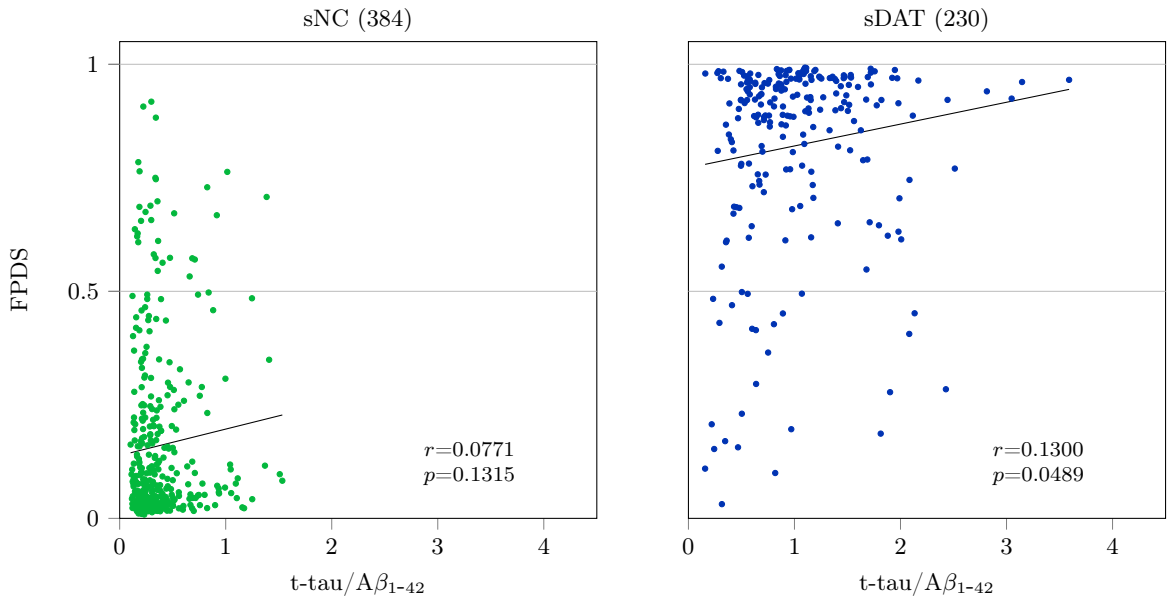
\begin{figure}
  \centering
  \caption{Pearson correlation between CSF \TauAbetaRatio\ and \FDGPETscoreShort\ across the sNC and sDAT images (baseline and follow-up combined). The CSF \TauAbetaRatio\ measures were only available for a subset of images and their numbers are shown in parentheses. The statistical significance threshold for correlation coefficient ($r$) was set at $p$<$0.05$.}
  \subfloat{\begin{tikzpicture}
\begin{axis}[title={sNC (384)},title style={font=\small,align=center,yshift=-6pt},clip=false,
width=0.48\linewidth,height=0.48\linewidth,point meta=explicit symbolic,scatter,
scatter/classes={sNC={mark=*,draw=sNC,fill=sNC,opacity=1,mark size=1}},
xmin=0.0000,xmax=4.5000,
xtick={0,1,2,3,4},
xticklabels={0,1,2,3,4},
ymin=0.0000,ymax=1.0500,
ytick={0,0.5,1},
yticklabels={0,0.5,1},
extra x ticks={},extra y ticks={0.5,1},
extra x tick labels = {},extra y tick labels = {},extra tick style = {grid=major},
xlabel={\TauAbetaRatio},xlabel shift = 0pt,
ylabel={\FDGPETscoreShort},ylabel shift = 0pt,
x tick label style={font=\small},y tick label style={font=\small},x label style={font=\small},y label style={font=\small},
xtick pos = left,xtick align = outside,
ytick pos = left,ytick align = outside
]
\addplot [color=white] table [x=x,y=y,meta=meta,only marks] {corr-sNC.txt};
\addplot [no markers,sharp plot,color=black,samples=2,domain=0.1029:1.5340] {(0.0582 * x) + (0.1384)};
\node [align=center,font=\footnotesize] at (axis cs:3.3,0.12) {$r$=$0.0771$\\$p$=$0.1315$};
\end{axis}
\end{tikzpicture}
}\qquad
  \subfloat{\begin{tikzpicture}
\begin{axis}[title={sDAT (230)},title style={font=\small,align=center,yshift=-6pt},clip=false,
width=0.48\linewidth,height=0.48\linewidth,point meta=explicit symbolic,scatter,
scatter/classes={sDAT={mark=*,draw=sDAT,fill=sDAT,opacity=1,mark size=1}},
xmin=0.0000,xmax=4.5000,
xtick={0,1,2,3,4},
xticklabels={0,1,2,3,4},
ymin=0.0000,ymax=1.0500,
ytick=\empty,
ytick style = {draw=none},
extra x ticks={},extra y ticks={0.5,1},
extra x tick labels = {},extra y tick labels = {},extra tick style = {grid=major},
xlabel={\TauAbetaRatio},xlabel shift = 0pt,
x tick label style={font=\small},y tick label style={font=\small},x label style={font=\small},y label style={font=\small},
xtick pos = left,xtick align = outside,
ytick pos = left,ytick align = outside
]
\addplot [color=white] table [x=x,y=y,meta=meta,only marks] {corr-sDAT.txt};
\addplot [no markers,sharp plot,color=black,samples=2,domain=0.1573:3.5902] {(0.0481 * x) + (0.7721)};
\node [align=center,font=\footnotesize] at (axis cs:3.3,0.12) {$r$=$0.1300$\\$p$=$0.0489$};
\end{axis}
\end{tikzpicture}}
\label{fig:correlation_plots-training}
\end{figure}

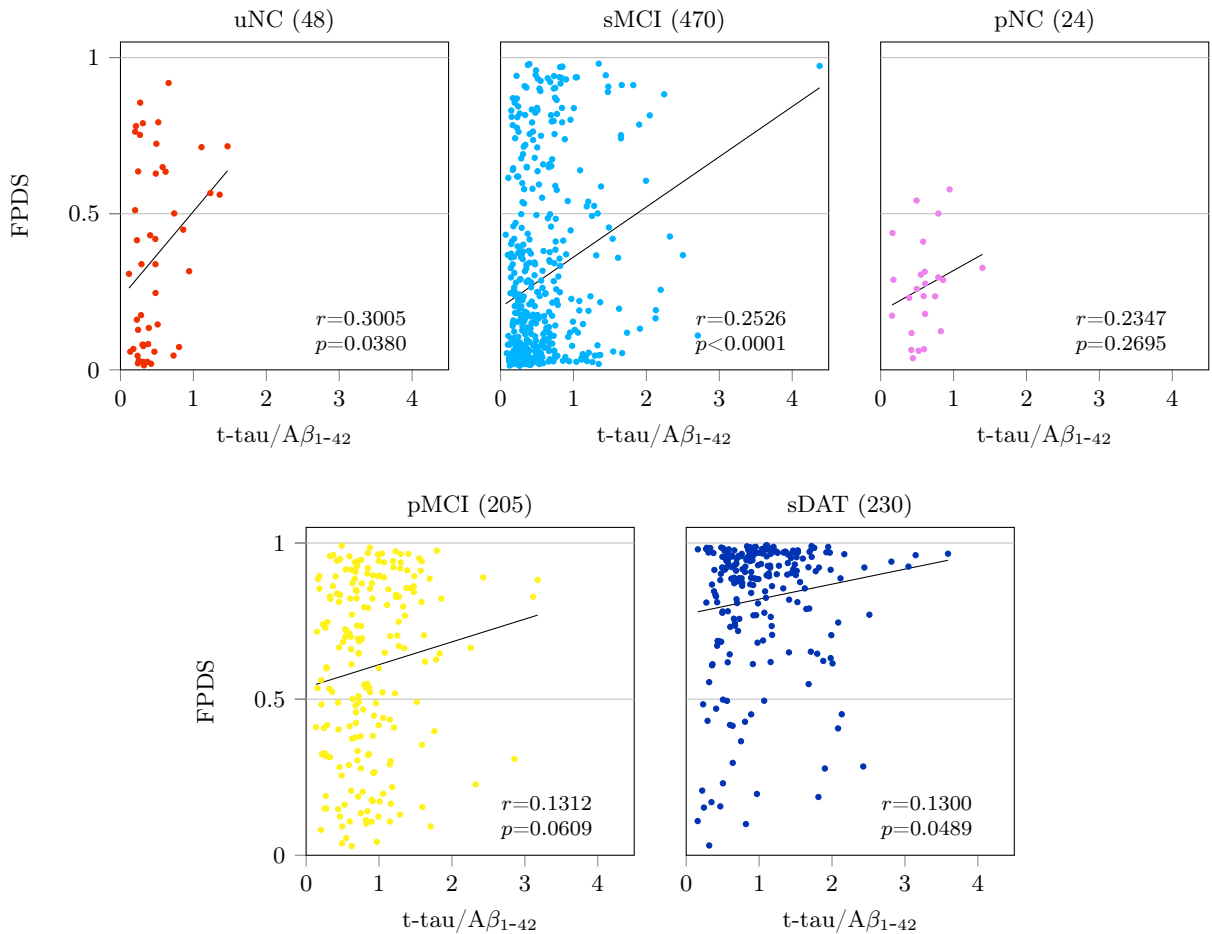
\begin{figure*}
  \centering
  \caption{Pearson correlation between the CSF \TauAbetaRatio\ and \FDGPETscoreShort\ values across the independent validation image groups. Only a subset of images in each group had CSF \TauAbetaRatio\ measures available. The number of images with \TauAbetaRatio\ values are given in parentheses. Correlation coefficient ($r$) was considered significant at $p$<$0.05$.}
  \subfloat{\begin{tikzpicture}
\begin{axis}[title={uNC (48)},title style={font=\small,align=center,yshift=-6pt},clip=false,
width=0.36\linewidth,height=0.36\linewidth,point meta=explicit symbolic,scatter,
scatter/classes={uNC={mark=*,draw=uNC,fill=uNC,opacity=1,mark size=1}},
xmin=0.0000,xmax=4.5000,
xtick={0,1,2,3,4},
xticklabels={0,1,2,3,4},
ymin=0.0000,ymax=1.0500,
ytick={0,0.5,1},
yticklabels={0,0.5,1},
extra x ticks={},extra y ticks={0.5,1},
extra x tick labels = {},extra y tick labels = {},extra tick style = {grid=major},
xlabel={\TauAbetaRatio},xlabel shift = 0pt,
ylabel={\FDGPETscoreShort},ylabel shift = 0pt,
x tick label style={font=\small},y tick label style={font=\small},x label style={font=\small},y label style={font=\small},
xtick pos = left,xtick align = outside,
ytick pos = left,ytick align = outside
]
\addplot [color=white] table [x=x,y=y,meta=meta,only marks] {corr-uNC.txt};
\addplot [no markers,sharp plot,color=black,samples=2,domain=0.1172:1.4683] {(0.2781 * x) + (0.2292)};
\node [align=center,font=\footnotesize] at (axis cs:3.3,0.12) {$r$=$0.3005$\\$p$=$0.0380$};
\end{axis}
\end{tikzpicture}
}\,
  \subfloat{\begin{tikzpicture}
\begin{axis}[title={sMCI (470)},title style={font=\small,align=center,yshift=-6pt},clip=false,
width=0.36\linewidth,height=0.36\linewidth,point meta=explicit symbolic,scatter,
scatter/classes={sMCI={mark=*,draw=sMCI,fill=sMCI,opacity=1,mark size=1}},
xmin=0.0000,xmax=4.5000,
xtick={0,1,2,3,4},
xticklabels={0,1,2,3,4},
ymin=0.0000,ymax=1.0500,
ytick=\empty,
ytick style = {draw=none},
extra x ticks={},extra y ticks={0.5,1},
extra x tick labels = {},extra y tick labels = {},extra tick style = {grid=major},
xlabel={\TauAbetaRatio},xlabel shift = 0pt,
x tick label style={font=\small},y tick label style={font=\small},x label style={font=\small},y label style={font=\small},
xtick pos = left,xtick align = outside,
ytick pos = left,ytick align = outside
]
\addplot [color=white] table [x=x,y=y,meta=meta,only marks] {corr-sMCI.txt};
\addplot [no markers,sharp plot,color=black,samples=2,domain=0.0704:4.3760] {(0.1607 * x) + (0.2003)};
\node [align=center,font=\footnotesize] at (axis cs:3.3,0.12) {$r$=$0.2526$\\$p$<$0.0001$};
\end{axis}
\end{tikzpicture}
}\,
  \subfloat{\begin{tikzpicture}
\begin{axis}[title={pNC (24)},title style={font=\small,align=center,yshift=-6pt},clip=false,
width=0.36\linewidth,height=0.36\linewidth,point meta=explicit symbolic,scatter,
scatter/classes={pNC={mark=*,draw=pNC,fill=pNC,opacity=1,mark size=1}},
xmin=0.0000,xmax=4.5000,
xtick={0,1,2,3,4},
xticklabels={0,1,2,3,4},
ymin=0.0000,ymax=1.0500,
ytick=\empty,
ytick style = {draw=none},
extra x ticks={},extra y ticks={0.5,1},
extra x tick labels = {},extra y tick labels = {},extra tick style = {grid=major},
xlabel={\TauAbetaRatio},xlabel shift = 0pt,
x tick label style={font=\small},y tick label style={font=\small},x label style={font=\small},y label style={font=\small},
xtick pos = left,xtick align = outside,
ytick pos = left,ytick align = outside
]
\addplot [color=white] table [x=x,y=y,meta=meta,only marks] {corr-pNC.txt};
\addplot [no markers,sharp plot,color=black,samples=2,domain=0.1569:1.3967] {(0.1308 * x) + (0.1873)};
\node [align=center,font=\footnotesize] at (axis cs:3.3,0.12) {$r$=$0.2347$\\$p$=$0.2695$};
\end{axis}
\end{tikzpicture}}\\
  \subfloat{\begin{tikzpicture}
\begin{axis}[title={pMCI (205)},title style={font=\small,align=center,yshift=-6pt},clip=false,
width=0.36\linewidth,height=0.36\linewidth,point meta=explicit symbolic,scatter,
scatter/classes={pMCI={mark=*,draw=pMCI,fill=pMCI,opacity=1,mark size=1}},
xmin=0.0000,xmax=4.5000,
xtick={0,1,2,3,4},
xticklabels={0,1,2,3,4},
ymin=0.0000,ymax=1.0500,
ytick={0,0.5,1},
extra x ticks={},extra y ticks={0.5,1},
extra x tick labels = {},extra y tick labels = {},extra tick style = {grid=major},
xlabel={\TauAbetaRatio},xlabel shift = 0pt,
ylabel={\FDGPETscoreShort},ylabel shift = 0pt,
x tick label style={font=\small},y tick label style={font=\small},x label style={font=\small},y label style={font=\small},
xtick pos = left,xtick align = outside,
ytick pos = left,ytick align = outside
]
\addplot [color=white] table [x=x,y=y,meta=meta,only marks] {corr-pMCI.txt};
\addplot [no markers,sharp plot,color=black,samples=2,domain=0.1354:3.1751] {(0.0729 * x) + (0.5376)};
\node [align=center,font=\footnotesize] at (axis cs:3.3,0.12) {$r$=$0.1312$\\$p$=$0.0609$};
\end{axis}
\end{tikzpicture}
}\,
  \subfloat{\begin{tikzpicture}
\begin{axis}[title={sDAT (230)},title style={font=\small,align=center,yshift=-6pt},clip=false,
width=0.36\linewidth,height=0.36\linewidth,point meta=explicit symbolic,scatter,
scatter/classes={sDAT={mark=*,draw=sDAT,fill=sDAT,opacity=1,mark size=1}},
xmin=0.0000,xmax=4.5000,
xtick={0,1,2,3,4},
xticklabels={0,1,2,3,4},
ymin=0.0000,ymax=1.0500,
ytick=\empty,
ytick style = {draw=none},
extra x ticks={},extra y ticks={0.5,1},
extra x tick labels = {},extra y tick labels = {},extra tick style = {grid=major},
xlabel={\TauAbetaRatio},xlabel shift = 0pt,
x tick label style={font=\small},y tick label style={font=\small},x label style={font=\small},y label style={font=\small},
xtick pos = left,xtick align = outside,
ytick pos = left,ytick align = outside
]
\addplot [color=white] table [x=x,y=y,meta=meta,only marks] {corr-sDAT.txt};
\addplot [no markers,sharp plot,color=black,samples=2,domain=0.1573:3.5902] {(0.0481 * x) + (0.7721)};
\node [align=center,font=\footnotesize] at (axis cs:3.3,0.12) {$r$=$0.1300$\\$p$=$0.0489$};
\end{axis}
\end{tikzpicture}
}
\label{fig:correlation_plots-validation}
\end{figure*}

To investigate the causal association of \FDGPETscoreShort\ with established ADP measures, Pearson correlation analysis was performed between the CSF \TauAbetaRatio\ and \FDGPETscoreShort\ values. The correlation results obtained across the training (Figure \ref{fig:correlation_plots-training}) and independent validation (Figure \ref{fig:correlation_plots-validation}) image groups are reported. In the case of training image groups, correlation analysis was performed using the combined set of baseline and follow-up images in each of the sNC and sDAT groups respectively. The spread of \TauAbetaRatio\ values in the sNC group (0.1 - 1.54) was relatively narrow as compared to sDAT group (0.15 - 3.6). Both the sNC and sDAT groups showed a weak yet positive correlation between the \TauAbetaRatio\ and \FDGPETscoreShort, with the sDAT group showing a relatively stronger correlation coefficient ($r$=0.13) that was also statistically significant ($p$=0.0489). Among the validation image groups, the \TauAbetaRatio\ values of the NC groups (uNC and pNC) had a relatively narrow range (0.11 - 1.47) compared to the other DAT- (sMCI) and DAT+ (pMCI and eDAT) groups. In general, the \FDGPETscoreShort\ was weakly, but positively correlated with \TauAbetaRatio. The correlation coefficient ranged between 0.13 - 0.31 among the various groups considered. However, correlation coefficients exhibited by DAT- groups (uNC and sMCI) were found to be statistically significant ($p$=0.0380 and $p$<0.0001), whereas the DAT+ (pNC, pMCI and eDAT) groups only exhibited a trend of positive correlations with $r$-values in the 0.13 - 0.24 interval.

\section{Discussion}
In this paper we report the development and validation of a novel \FDGPETscore\ (\FDGPETscoreShort). We computed the \FDGPETscoreShort\ using a multi-scale supervised ensemble learning approach on FDG-PET images. The \FDGPETscoreShort\ is a single scalar value between 0 and 1. It indicates the probability of the brain metabolism profile captured in a subject's FDG-PET image to be belonging to the DAT+ trajectory. The \FDGPETscoreShort\ was developed in an ensemble machine-learning paradigm trained on FDG-PET images belonging to sNC and sDAT subjects from the ADNI database. \FDGPETscoreShort\ as a DAT biomarker was then comprehensively validated on a large number of ADNI FDG-PET images ($N$=$2984$) across the sNC, uNC, sMCI, pNC, pMCI, eDAT and sDAT stratification.

\subsection{Real-world stratification scheme} The proposed stratification of imaging data into the 7 groups (Table \ref{tab:group_definitions_demographics}) provided a clinically relevant perspective for the development of the \FDGPETscoreShort\ framework. Particularly, the stratification scheme helped establish a clear delineation between images taken from subjects on the DAT- and DAT+ trajectories, and thus formulating DAT biomarker discovery as a supervised machine learning problem of building a classification model that can predict the probability of an image belonging to either the DAT- or DAT+ class. Most previous studies on imaging biomarkers were limited to stratifying images based on the NC, MCI and DAT diagnostic labels assigned at the time of image acquisition \cite{Rathore2017}. However, in recent years there has been interest in developing early stage DAT biomarkers adopting a sMCI/pMCI stratification of images associated with a clinical diagnosis of MCI \cite{Tong2017}. Our novel approach extends this MCI image stratification idea to the entire DAT spectrum by also stratifying the NC and DAT images into the sNC/uNC/pNC and eDAT/sDAT groups respectively. This enabled the validation of the \FDGPETscoreShort\ in a realistic experimental setting that is quite close to a practical clinical setup, where images from the uNC, sMCI, pNC, pMCI and eDAT groups were completely blinded from the trained ensemble classification model. We put forth our stratification approach as an ideal benchmark to evaluate future DAT biomarker methods. 

\subsection{Characteristics of the stratified groups} The training and validation image sets used in our analysis were found to be unbiased with respect to the associated relevant non-imaging phenotypic information, justifying the ignoring of non-imaging covariates in the proposed supervised learning framework. The age, MMSE and CSF \TauAbetaRatio\ values observed across the stratified groups did not reveal any anomalous group difference patterns (Table \ref{tab:group_definitions_demographics} and Table \ref{tab:pvalue_demographics}) that could potentially confound the proposed FDG-PET imaging based analysis. Most importantly, the sNC and sDAT groups used for training the \FDGPETscoreShort\ model had similar mean ages, and further as expected the sNC group had a significantly higher MMSE but a significantly lower \TauAbetaRatio\ compared to the sDAT group. Moreover, mean ages among the sMCI, pMCI and eDAT groups in the validation image set were also comparable to the training groups. The other two validation groups namely uNC and pNC had slightly, yet statistically significantly higher mean ages ($\sim$ 3 years older) than the training groups. However, this significant group difference might just be reflective of a sampling bias given that the uNC ($N$=110) and pNC ($N$=58) groups have considerably fewer images compared to the training groups, sNC ($N$=753) and sDAT ($N$=464). The group differences in MMSE and \TauAbetaRatio\ values between the validation and training groups followed known patterns, where the DAT- groups (uNC and sMCI) showed significantly higher mean MMSE but significantly lower mean \TauAbetaRatio\ when compared to the sDAT group, whereas the DAT+ groups (pNC, pMCI and eDAT) had significantly lower mean MMSE but significantly higher mean \TauAbetaRatio\ in comparison to the sNC group.

\subsection{\FDGPETscoreShort\ computation model characteristics} Two aspects of the trained ensemble classification model warrant further discussion, viz., the ROIs chosen by the model for \FDGPETscoreShort\ computation (Table \ref{tab:discrimROIs}), and the model's predictive performance on the sNC and sDAT groups (Figure \ref{fig:fpds_of_sNC_sDAT}). 

The ROIs selected by the ensemble model included parieto-temporal regions along with precuneus and cingulate gyrus. Recent studies have demonstrated the hypometabolism of parieto-temporal regions including precuneus and posterior cingluate as earliest evidences for MCI progression to DAT \cite{Arbizu2013, Ewers2014}. Further, left hemisphere regions were chosen more often compared to their corresponding contralateral regions. Similar, preferential left sided hypometabolism was reported during early stages of DAT \cite{Brown2014}. Hence, the ROIs chosen by the ensemble model for \FDGPETscoreShort\ computation are consistent with established spatial hypometabolism patterns in DAT.

The ensemble model's \FDGPETscoreShort\ predictions on the sNC and sDAT images were consistent with the fact that these images belong to individuals who are at the extremities of the DAT spectrum, i.e., the \FDGPETscoreShort\ distributions of sNC and sDAT were skewed and only had a small overlap with AUCs of 0.95 and 0.98 for the baseline and follow-up subgroups respectively. Interestingly, among the sNC images with \FDGPETscoreShort>0.5 (misclassified as DAT+), the mean \TauAbetaRatio\ was found to be slightly higher (0.42 vs 0.36, $p$=0.2463) than the sNC images with \FDGPETscoreShort<=0.5 (correctly identified as DAT-). Whereas, in sDAT images with \FDGPETscoreShort<=0.5 (misclassified as DAT-), the mean \TauAbetaRatio\ was statistically significantly lower (0.84 vs 1.06, $p$=0.0073) when compared to sDAT images with \FDGPETscoreShort>0.5 (accurately labeled as DAT+). These observations agree with the positive correlations found between the \TauAbetaRatio\ and \FDGPETscoreShort\ values among the sNC and sDAT groups respectively (Figure \ref{fig:correlation_plots-training}). The occurrence of DAT like metabolism patterns (higher \FDGPETscoreShort) among the misclassified sNC might be owing to their increased \TauAbetaRatio\ values and in a similar manner the shift away from DAT metabolism patterns (lower \FDGPETscoreShort) among the misclassified sDAT could be attributed to their relatively lower \TauAbetaRatio\ values.

The predictive performance of the ensemble model is on par with (or better than) the sNC vs sDAT classification results published in latest FDG-PET imaging based studies, which showed AUCs ranging from 0.93 on a cohort of 52 sNC and 51 sDAT images \cite{Ye2016} to 0.97 on a 117 sNC and 113 sDAT cohort \cite{Li2017}. These studies evaluated their classification models using a 10-fold cross-validation scheme, which is known \cite{Bylander2002} to produce generalization error estimates (measure of predictive performance on unseen data) similar to that of the out-of-bag prediction scheme, that was used for evaluation of the ensemble model on baseline sNC and sDAT images. However, arguably the ensemble model's performance on the follow-up images gives a much better estimate of the generalization error, as the follow-up images were completely hidden during the ensemble model training process and hence can be considered as unseen data, despite their implicit relation to the corresponding baseline images. It's also important to highlight the relatively large sample size of the follow-up image set (393 sNC and 226 sDAT images) in comparison to the cohorts used in previous FDG-PET studies \cite{Ye2016, Li2017, Weiner2017}. This further underscores the confidence in the reported predictive performance of the ensemble model on sNC and sDAT groups.

\subsection{Comprehensive evaluation of the \FDGPETscoreShort\ computation model} The ensemble classification model's predictive performance evaluated on a large independent validation set of images ($N$=1767), taken from individuals at different stages of AD spectrum, provided a rigorous and a realistic way to assess the potential of using \FDGPETscoreShort\ for DAT diagnosis (Figure \ref{fig:fpds_of_uNC_sMCI_pNC_pMCI_eDAT}). The ensemble model achieved an AUC of 0.78 in discriminating the DAT- (uNC and sMCI) and the DAT+ (pNC, pMCI and eDAT) groups, strongly advocating the consideration of \FDGPETscoreShort\ as a DAT biomarker. A more detailed analysis of the \FDGPETscoreShort\ predictions across the DAT- and DAT+ groups revealed a non-trivial association between the \TauAbetaRatio\ and \FDGPETscoreShort\ values. The DAT- images with \FDGPETscoreShort>0.5 (misclassified as DAT+) had statistically significantly higher mean \TauAbetaRatio\ (0.71 vs 0.49, $p$<0.0001) compared to the DAT- images with \FDGPETscoreShort<=0.5 (correctly labeled as DAT-). While, the mean \TauAbetaRatio\ among the DAT+ images with \FDGPETscoreShort<=0.5 (misclassified as DAT-) was found to be significantly lower (0.74 vs 0.94, $p$=0.0015) compared to the DAT+ images with \FDGPETscoreShort>0.5 (correctly labeled as DAT+). In light of these findings, along with the positive correlations observed between \TauAbetaRatio\ and \FDGPETscoreShort\ within each of the DAT- and DAT+ groups (Figure \ref{fig:correlation_plots-validation}), it can be speculated that the relatively higher \TauAbetaRatio\ values might be triggering the presence of DAT like metabolism patterns (higher \FDGPETscoreShort) in misclassified DAT-. Similarly, the comparatively lower \TauAbetaRatio\ values could be the underlying cause behind the lack of DAT like metabolism patterns (lower \FDGPETscoreShort) among the misclassified DAT+.

The predicted \FDGPETscoreShort\ values for the sMCI images were observed to increase with age (Figure \ref{fig:fpds_vs_AGE}), i.e., images corresponding to older subjects tended to have higher \FDGPETscoreShort\ values compared to images taken from younger subjects. In particular, when comparing the two subgroups of sMCI images whose age ranges were above and below the average sMCI age of $\sim$ 75 years (Table \ref{tab:group_definitions_demographics}) respectively, the mean \FDGPETscoreShort\ for the images in the older subgroup was found to be significantly greater than the mean \FDGPETscoreShort\ among the images from the younger subgroup (0.45 vs 0.22, $p$<0.0001). Further, as could be expected based on the statistically significant positive correlation observed between \FDGPETscoreShort\ and \TauAbetaRatio\ in sMCI ($r=0.2526$ with $p$<0.0001, Figure \ref{fig:correlation_plots-validation}), the older subgroup also showed a significantly higher mean \TauAbetaRatio\ compared to the younger subgroup (0.61 vs 0.51, $p$<0.0001). Apart from the sMCI group, none of the other uNC, pNC, pMCI and eDAT groups displayed any apparent age specific \FDGPETscoreShort\ patterns (Figure \ref{fig:fpds_vs_AGE}). Nevertheless, among these groups a trend of positive correlations between \TauAbetaRatio\ and \FDGPETscoreShort\ was observed (Figure \ref{fig:correlation_plots-validation}). This suggests a possible age independent causal relationship between \TauAbetaRatio\ and the occurrence of DAT like metabolism patterns.

In the pMCI group, the predicted \FDGPETscoreShort\ values were found to decrease with longer time to conversion (Figure \ref{fig:fpds_vs_TimeToConversion}). Notably, the mean \FDGPETscoreShort\ for the subgroup of images that were within 4 years to conversion was significantly higher than that of the image subgroup whose conversion times exceeded 4 years (0.67 vs 0.43, $p$<0.0001). Moreover, in concordance with the positive correlation observed between \TauAbetaRatio\ and \FDGPETscoreShort\ among the pMCI ($r=0.1312$ with $p$<0.0609, Figure \ref{fig:correlation_plots-validation}), the mean \TauAbetaRatio\ for the within 4 years to conversion subgroup was also significantly higher compared to the subgroup with longer than 4 year conversion times (0.91 vs 0.66, $p$=0.0229). Based on these findings it is conceivable that, from around 4 years prior to a clinical diagnosis of DAT there might be a noticeable increase in the \TauAbetaRatio\ causing a prevalence of DAT like metabolism patterns among the pMCI. While beyond the 4 year conversion window, it can be expected that there would be a considerable reduction in the appearance of DAT like metabolism patterns in pMCI. In fact, metabolic disruptions in an earlier NC stage of pMCI were found to be virtually undetectable as evidenced by the significantly lower mean \FDGPETscoreShort\ of the pNC group compared to the pMCI (0.35 vs 0.63, $p$<0.0001) and also indicated by the extremely low classification accuracy (0.28, Figure \ref{fig:fpds_vs_TimeToConversion}) achieved on pNC images.

\subsection{MCI conversion prediction - comparison with state-of-the-art} Several FDG-PET image analysis methods have previously been considered for addressing the task of predicting MCI to DAT conversion \cite{Young2013a,Zhu2014a,Cheng2015,Cheng2015a,Lange2015,Wang2016,Pagani2017,Inui2017,Liu2017a}. In these methods, the main idea is to train a binary classification model for separating the MCI into two groups, the sMCI which remain stable and the pMCI that convert to DAT in the future. Aside from the standard approach of using images from the sMCI and pMCI groups as training data \cite{Wang2016,Pagani2017,Zhu2014a}, some of the methods have augmented the training process with information derived from the sNC and sDAT images as well \cite{Liu2017a,Lange2015,Cheng2015,Cheng2015a}. Further, akin to the proposed approach for training the \FDGPETscoreShort\ computation model, there were a few methods that solely employed the sNC and sDAT images during the training phase \cite{Young2013a,Inui2017}. 

In Table \ref{tab:compar_with_MCI_conversion_state_of_the_art}, the sMCI vs pMCI classification results reported in the aforementioned works are summarized. Table \ref{tab:compar_with_MCI_conversion_state_of_the_art} also shows the AUC achieved when using the \FDGPETscoreShort\ to discriminate between the sMCI and the pMCI that are within 2, 3 and 5 years to conversion respectively. The proposed \FDGPETscoreShort\ based approach outperformed almost all of the state-of-the-art methods achieving an AUC of more than 0.77 in each of the three time to conversion cases. Only \cite{Wang2016} and \cite{Pagani2017} reported higher AUCs than the proposed approach. However, these two methods reported the cross-validated AUC, whereas a more challenging independent validation experiment was used to evaluate the performance of the \FDGPETscoreShort\ approach. Moreover, in \cite{Wang2016} and \cite{Pagani2017}, the classification model parameter tuning was done using the testing subsets of the cross-validation splits which leads to inflated estimates of the classification performance. In fact to avoid such an optimistic performance evaluation, the other methods reporting cross-validated AUCs \cite{Zhu2014a,Cheng2015,Cheng2015a} used ``nested'' cross-validation, where the classification model parameters were tuned on the training subsets of the cross-validation splits rather than the testing subsets. Last, it should be highlighted that the better performance of the \FDGPETscoreShort\ approach was demonstrated on a considerably larger sample size (>5x more images) compared to the other methods. 
\begin{table}
\centering
\caption{Comparison of sMCI vs pMCI classification performance obtained using \FDGPETscoreShort\ with the state-of-the-art FDG-PET based methods.}
\setlength{\tabcolsep}{0.06in}\begin{tabular}{*{5}{c}}
\hline
                    & sMCI:pMCI       &  Time to           & Evaluation                     &     \\
Study               & [images]        &  conversion        & scheme                         & AUC \\
\hline
\cite{Zhu2014a}     &   56:43         &        0-2 years     & 10-fold cross-validation       & 0.774\\
\cite{Cheng2015}    &   56:43         &        0-2 years     & 10-fold cross-validation       & 0.734\\
\cite{Cheng2015a}   &   56:43         &        0-2 years     & 10-fold cross-validation       & 0.741\\
\textbf{\FDGPETscoreShort}   &  \textbf{881:254}           &        \textbf{0-2 years}     & \textbf{independent validation}         & \textbf{0.806}      \\
\hline
\cite{Young2013a}   &   96:47         &        0-3 years     & independent validation         & 0.767\\
\cite{Lange2015}    &   181:60        &        0-3 years     & independent validation         & 0.746\\
\cite{Wang2016}     &   65:64         &        0-3 years     & leave-one-out cross-validation & 0.802\\
\cite{Liu2017a}     &  108:126        &        0-3 years     & prediction on training set          & 0.736\\
\textbf{\FDGPETscoreShort}   &  \textbf{881:362}           &        \textbf{0-3 years}     & \textbf{independent validation}         & \textbf{0.796}      \\
\hline
\cite{Pagani2017}   &   27:95         &        0-5 years     & 21-fold cross-validation       & 0.911\\
\cite{Inui2017}     &   19:49         &        0-5 years     & independent validation         & 0.712\\
\textbf{\FDGPETscoreShort}   &  \textbf{881:442}           &        \textbf{0-5 years}     & \textbf{independent validation}         & \textbf{0.772} \\
\end{tabular}



\label{tab:compar_with_MCI_conversion_state_of_the_art}
\end{table}

\subsection{Limitations and future directions} The results reported in this paper are understandably limited by the ADNI data characteristics. In general, the final (at time of death) clinical diagnosis for ADNI subjects is not known, this is because either the subjects are surviving or they were not followed-up around their demise. Consequently, it is possible that some subjects currently determined to be on the DAT- trajectory, i.e., with images belonging to the sNC/uNC/sMCI groups, might receive a clinical diagnosis of DAT in the future. While this is a limitation in our current results, in case the final diagnosis for such subjects becomes available, it would be interesting to review if the ensemble model had actually correctly predicted the sNC/uNC/sMCI images of these subjects as belonging to the DAT+ trajectory (\FDGPETscoreShort>0.5). Another limitation of the reported results, as mentioned before, is that the correlation analysis between the \FDGPETscoreShort\ and \TauAbetaRatio\ values was reported only on a subset of the images owing to partial availability of CSF measures in the ADNI database. In spite of these ADNI data related limitations, it is important to note that both the novel stratification scheme and the ensemble classification framework proposed in our work have a more general applicability and are not specific to the ADNI cohort used in this study. In fact, as part of future work we plan to extend our methodology to incorporate multimodal imaging data and validate it on other relevant AD neuroimaging databases.

\section{Acknowledgements}
Funding for this research is gratefully acknowledged from National Science Engineering Research Council (NSERC), Canadian Institutes of Health Research (CIHR), Brain Canada, Pacific Alzheimer's Research Foundation, the Michael Smith Foundation for Health Research (MSFHR), and the National Institute on Aging (R01 AG055121-01A1). We thank Compute Canada for the computational infrastructure provided for the data processing in this study. Data collection and sharing for this project was funded by the Alzheimer's Disease Neuroimaging Initiative
(ADNI) (National Institutes of Health Grant U01 AG024904) and DOD ADNI (Department of Defense award
number W81XWH-12-2-0012). ADNI is funded by the National Institute on Aging, the National Institute of
Biomedical Imaging and Bioengineering, and through generous contributions from the following: AbbVie,
Alzheimer’s Association; Alzheimer’s Drug Discovery Foundation; Araclon Biotech; BioClinica, Inc.; Biogen;
Bristol-Myers Squibb Company; CereSpir, Inc.; Cogstate; Eisai Inc.; Elan Pharmaceuticals, Inc.; Eli Lilly and
Company; EuroImmun; F. Hoffmann-La Roche Ltd and its affiliated company Genentech, Inc.; Fujirebio; GE
Healthcare; IXICO Ltd.; Janssen Alzheimer Immunotherapy Research \& Development, LLC.; Johnson \&
Johnson Pharmaceutical Research \& Development LLC.; Lumosity; Lundbeck; Merck \& Co., Inc.; Meso
Scale Diagnostics, LLC.; NeuroRx Research; Neurotrack Technologies; Novartis Pharmaceuticals
Corporation; Pfizer Inc.; Piramal Imaging; Servier; Takeda Pharmaceutical Company; and Transition
Therapeutics. The Canadian Institutes of Health Research is providing funds to support ADNI clinical sites
in Canada. Private sector contributions are facilitated by the Foundation for the National Institutes of Health
(www.fnih.org). The grantee organization is the Northern California Institute for Research and Education,
and the study is coordinated by the Alzheimer’s Therapeutic Research Institute at the University of Southern
California. ADNI data are disseminated by the Laboratory for Neuro Imaging at the University of Southern
California.

\clearpage
\addcontentsline{toc}{section}{References}
\bibliographystyle{unsrt}
\bibliography{references,mendeley}

\end{document}